\documentclass[DTMColor]{ROB-New}
\usepackage{siunitx}
\usepackage{tcolorbox}

\begin{document}

\authormark{Cambridge Authors}

\articletype{RESEARCH ARTICLE}

\jnlPage{1}{1}
\jyear{2026}
\jdoi{10.1017/xxxxx}

\title{Iteratively Learning Muscle Memory for Legged Robots to Master Adaptive and High Precision Locomotion}

\author[]{Jing Cheng}
\author[]{Yasser G. Alqaham}
\author[]{Zhenyu Gan\hyperlink{corr}{*}}
\author[]{Amit K. Sanyal}

\address[]{Department of Mechanical and Aerospace Engineering, Syracuse University, Syracuse, US}

\address{\hypertarget{corr}{*}Corresponding Author: Zhenyu Gan. \email{zgan02@syr.edu}}

\received{xx xxx xxx}
\revised{xx xxx xxx}
\accepted{xx xxx xxx}


\abstract{This paper presents a scalable and adaptive control framework for legged robots that integrates \textbf{Iterative Learning Control (ILC)} with a biologically inspired \textbf{torque library (TL)}, analogous to muscle memory. The proposed method addresses key challenges in robotic locomotion, including accurate trajectory tracking under unmodeled dynamics and external disturbances. By leveraging the repetitive nature of periodic gaits and extending ILC to nonperiodic tasks, the framework enhances accuracy and generalization across diverse locomotion scenarios.
The control architecture is data-enabled, combining a physics-based model derived from hybrid-system trajectory optimization with real-time learning to compensate for model uncertainties and external disturbances. A central contribution is the development of a generalized TL that stores learned control profiles and enables rapid adaptation to changes in speed, terrain, and gravitational conditions—eliminating the need for repeated learning and significantly reducing online computation.
The approach is validated on the bipedal robot Cassie and the quadrupedal robot A1 through extensive simulations and hardware experiments. Results demonstrate that the proposed framework reduces joint tracking errors by up to 85\% within a few seconds and enables reliable execution of both periodic and nonperiodic gaits, including slope traversal and terrain adaptation. Compared to state-of-the-art whole-body controllers, the learned skills eliminate the need for online computation during execution and achieve control update rates exceeding 30$\times$ those of existing methods. These findings highlight the effectiveness of integrating ILC with torque memory as a highly data-efficient and practical solution for legged locomotion in unstructured and dynamic environments.
}

\maketitle

\section{Introduction}
Legged robots are uniquely equipped to navigate unstructured environments and perform complex locomotion tasks, making them invaluable for applications such as disaster response, planetary exploration, and hazardous environment inspection. 
Unlike wheeled or tracked robots, legged systems excel in challenging terrains, where adaptability to uneven ground, sudden disturbances, and dynamic transitions is critical. Examples such as the quadrupedal A1 \cite{unitree2020a1} and bipedal Cassie \cite{agilityrobotics_cassie} demonstrate significant progress in legged robotics. However, the realization of precise, adaptive, and computationally efficient locomotion remains an ongoing challenge.
The core challenge in controlling legged robots arises from their hybrid dynamics, which involve continuous motion during stance phases and discrete transitions between contact events. These nonlinear dynamics are sensitive to uncertainties, such as unmodeled forces, actuator limitations, terrain variations and environmental disturbances. Locomotion control requires not only addressing these challenges but also achieving real-time performance with limited computational resources.

%
\begin{figure}[tbp]
\centering
\includegraphics[width=0.5\columnwidth]{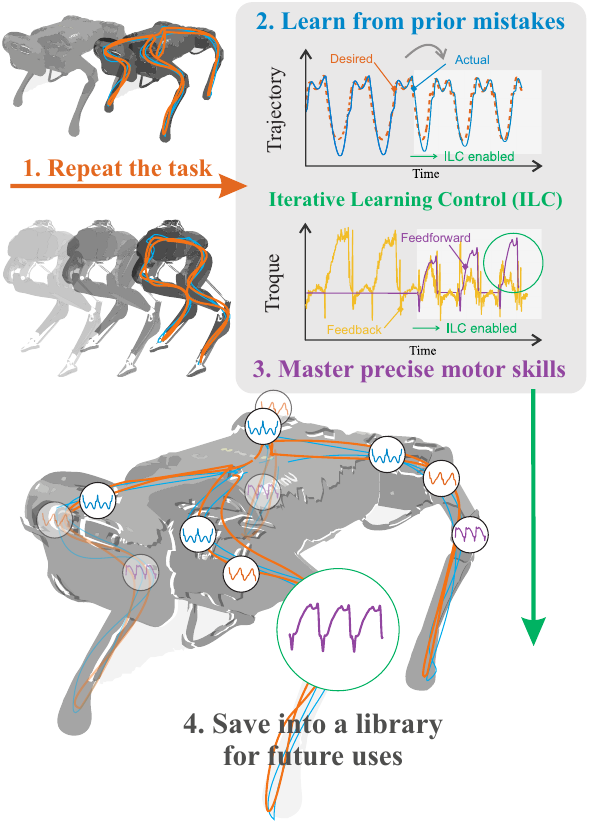}
\caption[Conceptual illustration of Iterative Learning Control (ILC)]{Conceptual illustration of the ILC framework. The ILC process iteratively refines the feedforward control inputs by leveraging data from previous iterations to minimize tracking errors. The figure highlights the interplay between feedback and feedforward components, showcasing how the control scheme adapts to improve trajectory tracking over successive iterations.}
\label{fig:ILC_Concept}
\vspace{-2mm}
\end{figure}
%
To address the challenges of legged locomotion, researchers have primarily adopted two paradigms: \textbf{model-based control methods} and \textbf{learning-based approaches}. Model Predictive Control (MPC) is widely used for its ability to optimize future trajectories while satisfying system constraints over finite horizons \cite{kim2019highly, Katz_2019}. Commonly, simplified models such as the Linear Inverted Pendulum (LIP) model \cite{kajita2001simple} or the centroidal dynamics model \cite{wieber2006trajectory} are used to reduce computational complexity and facilitate real-time planning. However, these models often oversimplify real robot dynamics and fail to capture unmodeled disturbances, actuator dynamics, or multi-contact scenarios. As a result, MPC's reliance on model fidelity and its computational intensity limit its scalability and robustness, particularly in unpredictable or high-speed dynamic settings \cite{neunert2018whole, kang2022nonlinearMPC}.
On the other hand, reinforcement learning (RL) offers a model-free paradigm that enables robots to learn control policies through environmental interaction. RL has demonstrated strong results in generating agile locomotion behaviors across various platforms, including ANYmal \cite{hwangbo2020learning} and Cassie \cite{siekmann2021blind}. Nevertheless, its widespread application faces several key challenges. These include data inefficiency \cite{gu2017deep, haarnoja2018soft}, requiring millions of samples to converge, and difficulties in transferring policies trained in simulation to the real world due to the sim-to-real gap \cite{peng2018sim, zhao2020sim}. Moreover, safety during exploration and policy degradation under distributional shift remain open issues \cite{hansen2021, yang2022RL}. These limitations motivate hybrid approaches that combine the strengths of physics-based control and data-driven learning.

To bridge this gap, we draw inspiration from biological motor learning, specifically the concept of muscle memory and propose a hybrid framework that integrates ILC with a generalized TL. ILC capitalizes on the repetitive structure of locomotion by refining control inputs based on tracking errors from previous iterations directly from the hardware under the influences of unmodeled dynamics and environmental disturbances \cite{ahn2007iterative, bristow2006survey}. Once a motion is learned, its feedforward control profile can be stored in a TL and recalled with minimal computation for future executions.
Figure~\ref{fig:ILC_Concept} illustrates this iterative process: feedback errors drive the refinement of feedforward inputs until convergence, at which point the learned torques can be preserved and reused. This enables rapid adaptation across speeds, terrains, and environmental conditions without repeated learning.
This work expands prior research \cite{cheng2023practice} by generalizing ILC and torque libraries to accommodate both periodic and nonperiodic motions. Key contributions are as follows:
\begin{itemize}
    \item \textbf{Hybrid Control Framework}: Integration of ILC and torque library idea in a hybrid system model to handle the discrete and continuous dynamics of legged locomotion, ensuring stability across diverse gaits and transitions.
    \item \textbf{Generalized Torque Libraries}: An expanded TL concept that supports nonperiodic tasks and enables robots to adapt to varying terrains, speeds, and gravitational conditions with minimal retraining.
    \item \textbf{Validation Across Platforms}: Experimental validation on the bipedal Cassie robot and quadrupedal A1 robot, demonstrating scalability and significant improvements in trajectory tracking, energy efficiency, and stability.
\end{itemize}

Our experimental and simulation results demonstrate that the proposed control framework substantially improves joint trajectory tracking for both the quadrupedal robot A1 and the bipedal robot Cassie. Specifically, the method achieves up to \textbf{85\% reduction in tracking errors within just a few seconds} of execution time. This rapid convergence highlights the data efficiency of the approach, as accurate feedforward compensation is learned without requiring large datasets or prolonged training.
The framework generalizes across platforms and supports a wide range of tasks, including slope traversal, terrain adaptation, and nonperiodic maneuvers, all without relying on online optimization or repeated learning. Once a locomotion skill is learned through a few repeated trials, it is stored in a TL and can be recalled instantly for future execution. This eliminates the need for computationally expensive model inversion or trajectory replanning at runtime.
Compared to state-of-the-art approaches, \textbf{our method achieves comparable or superior tracking performance while reducing runtime computation by more than an order of magnitude.} This decoupling of motion planning from execution enables the robot to adapt dynamically to disturbances and environmental variations, making the proposed control scheme a generalizable and scalable solution for real-world legged locomotion.

The remainder of this paper is organized as follows. Section~\ref{sec:related_work} surveys prior work in legged robot control, with an emphasis on both model-based strategies and data-driven learning approaches. Section~\ref{sec:methods} details the proposed hybrid control framework, including the integration of ILC and the construction of the TL. Section~\ref{sec:results} presents quantitative and qualitative results from extensive simulations and hardware experiments. Section~\ref{sec:discussion} discusses the broader implications, comparisons to existing methods, and limitations of the approach. Finally, Section~\ref{sec:conclusion} summarizes the contributions and outlines directions for future research.

\section{Related Work}
\label{sec:related_work}
Legged robot control has undergone significant advancements over the past decade, driven by the demand for versatile, adaptive, and robust locomotion systems. Two dominant approaches: model-based methods, such as MPC; and learning-based methods, including RL, have demonstrated considerable promise in addressing the challenges associated with the hybrid dynamics of legged robots. This section reviews the state of the art in these domains, outlining key advancements and limitations.

\subsection{Model-Based Methods}

Model-based methods synthesize control strategies by optimizing over predictive models of a robot’s dynamics. Among these, MPC has been widely adopted for its ability to compute control inputs that minimize a cost function over a finite horizon while satisfying physical and actuation constraints \cite{kim2019highly, Katz_2019, di2018dynamic}. To enable real-time implementation, MPC is typically paired with simplified representations such as the LIP or Single Rigid Body Dynamics (SRBD) \cite{9196562, chadwick2020vitruvio}, which abstract away full-body nonlinearities while preserving balance-critical dynamics.
These strategies have been successfully deployed on a variety of platforms. For example, the MIT Mini Cheetah demonstrated dynamic trotting and bounding using an MPC controller that optimized centroidal dynamics and translated the results to joint torques via inverse dynamics \cite{Katz_2019}. Similarly, Cassie and other bipedal robots have employed centroidal models to generate agile walking motions \cite{kim2019highly, da20162d}.

Nonetheless, a key limitation of these simplified models lies in their inability to capture joint-level dynamics, actuator constraints, and interaction forces such as friction, compliance, or impact effects. To address these shortcomings, MPC is often integrated with \textbf{Whole-Body Control (WBC)} to compute joint torques based on full-body dynamics once the desired ground reaction forces (GRFs) are generated. WBC solves a constrained inverse dynamics problem, providing accurate torque outputs that incorporate contact and hardware-specific constraints.
This hierarchical MPC+WBC structure provides enhanced fidelity and robustness but introduces trade-offs in control rate and predictive horizon. Due to solver limitations and the inherent approximation errors over long time horizons, these systems typically run at 200–300~Hz with prediction windows of about 1~second \cite{grandia2019feedback}. While the WBC back-end compensates for many modeling imperfections, the controller’s overall performance still depends on the simplifications made in the MPC stage and the precision of the robot model used in WBC.

To bypass these limitations, recent efforts have focused on \textbf{Nonlinear MPC (NMPC)} formulations that incorporate full-body dynamics directly into the optimization problem. Unlike SRBD-based MPC, NMPC explicitly models multibody kinematics, joint-level dynamics, and contact interactions, enabling higher accuracy and improved performance during contact-rich maneuvers or under external disturbances. For instance, Rathod et al.~\cite{rathod2021nmpc} developed a whole-body NMPC framework for HyQ that enabled omni-directional walking and terrain adaptation. Bratta et al.~\cite{bratta2021nmpc} introduced a mobility-aware NMPC criterion for navigating constrained environments. Other works have explored contact-complicit \cite{Le2024CIMPC}, soft contact modeling \cite{winkler2018gait}, and real-time implementation on humanoids \cite{neunert2018whole}.

Despite these advances, NMPC remains computationally intensive. Prediction horizons are often limited to under 2~seconds, and solver rates rarely exceed 25~Hz due to the high dimensionality and nonlinearity of the problem \cite{rathod2021nmpc}. These frameworks typically involve over 2000 decision variables and require hardware acceleration or parallelization to meet real-time constraints. Moreover, the accuracy of NMPC relies on precise inertial, contact, and actuator modeling, which can be challenging to obtain and maintain in practice.
In general, both MPC and NMPC provide powerful frameworks for legged locomotion. While MPC+WBC offers a modular structure and reasonable computational cost, it relies on upstream simplifications that may limit its adaptability. NMPC, on the other hand, eliminates model reduction but demands significantly more computation. Notably, both approaches incur high computational loads even during steady-state locomotion such as walking at constant speed or stepping in place. These limitations highlight the importance of lightweight, data-enabled alternatives that can achieve high control performance without constant re-solving of complex optimization problems.

\subsection{Learning-Based Methods}

Learning-based methods provide a model-free alternative to traditional control by enabling robots to acquire control policies directly from data. Among these, RL has demonstrated impressive capabilities across a range of locomotion tasks. Quadrupeds such as ANYmal have used RL to learn terrain-aware policies that generalize to unstructured environments with significant visual and physical variability \cite{hwangbo2020learning}. Similarly, Cassie has shown the ability to climb stairs and traverse obstacles without vision, relying on proprioceptive feedback alone \cite{siekmann2021blind}.
A major advancement in this domain has been the development of sim-to-real transfer techniques, which aim to bridge the discrepancy between simulation and physical systems. Domain randomization \cite{tobin2017domain} and dynamics randomization \cite{peng2018sim, tan2018sim} expose policies to a wide range of simulated conditions during training, promoting robustness to unmodeled dynamics and sensor noise. These strategies have enabled RL policies trained entirely in simulation to be successfully deployed on hardware, often without additional fine-tuning. For example, Tan et al. \cite{tan2018sim} demonstrated high-speed quadrupedal locomotion on real hardware after extensive randomization in simulation.

Despite their promise, RL methods face several key limitations. Most algorithms are extremely data-hungry, often requiring millions of environment interactions to achieve acceptable performance \cite{haarnoja2018soft, gu2017deep}. This leads to long training times and heavy reliance on simulation infrastructure, making it difficult to adapt policies online or in-the-loop on physical systems. Additionally, policies trained in simulation may fail in the real world due to subtle mismatches in dynamics, contact modeling, and sensor characteristics—a challenge often referred to as the sim-to-real gap \cite{zhao2020sim}.
Another fundamental drawback lies in safety and constraint satisfaction. Unlike model-based methods such as MPC, which enforce constraints explicitly during optimization, RL often lacks formal guarantees on stability, feasibility, or safety. This is particularly problematic during early-stage training, where exploratory behaviors can lead to unsafe joint configurations or hardware damage. While constrained RL \cite{achiam2017constrained, chow2018lyapunov} and safe exploration strategies \cite{hansen2021} have been proposed, they remain difficult to scale and often complicate the training process.

Furthermore, generalization beyond trained tasks or conditions remains an open challenge. RL policies typically overfit to specific training distributions and degrade under distributional shift. Recent efforts to improve generalization, such as hierarchical RL \cite{nachum2018data} and meta-learning \cite{finn2017model}, offer promising directions, but these approaches are still in early development and not yet widely adopted in real-world robotic systems.
While RL-based control has achieved impressive empirical results, its high sample complexity, training overhead, and lack of safety guarantees limit its practical deployment in many real-world legged locomotion scenarios. Bridging these gaps remains a critical area of research, motivating hybrid frameworks that combine the adaptability of learning with the structure and safety of physics-based control.

\subsection{Data-Augmented Control Strategies}

To overcome the limitations of purely model-based or purely learning-based techniques, data-augmented control strategies have emerged as a promising class of methods that enhance traditional controllers with data-driven components. These approaches preserve the structure, interpretability, and safety guarantees of physics-based models while leveraging data to improve performance, robustness, or generalization.
A common approach involves combining MPC with RL, where MPC handles long-horizon planning while RL compensates for unmodeled dynamics at the torque or policy level. For example,  Yang et al. \cite{yang2020data} demonstrated a hierarchical framework that integrates MPC with an RL-based low-level controller to improve robustness during terrain transitions. Similarly, residual learning has been applied to learn the difference between idealized models and real-world dynamics, enhancing controllers based on centroidal dynamics \cite{xie2021glide} or learning torque \cite {chen2023learningtorque}.

In this work, we adopt ILC as a structured and data-efficient alternative. ILC exploits the repetitive nature of tasks—such as pronking, trotting, or jumping—to iteratively refine feedforward control inputs. It operates directly in torque space and adjusts control commands based on tracking errors from previous executions, requiring no model of system uncertainty. Recent studies show that ILC significantly improves tracking performance in quadrupeds with minimal data \cite{cheng2023practice,10590932}.
We select ILC for its fast convergence, low data requirements, and robustness to unmodeled effects. Unlike RL, which often demands large datasets and extensive exploration, ILC achieves substantial improvements within a few repetitions. It also does not rely on a precise dynamic model, making it well-suited to handle modeling errors, actuator delays, and contact disturbances. Furthermore, its phase-indexed structure enables adaptation to a wide range of task parameters, including varying speeds, terrain conditions, and robot configurations. These properties make ILC a practical and powerful tool for enhancing trajectory tracking in agile and dynamic locomotion.

\section{Proposed Methods}
\label{sec:methods}

This section introduces the proposed data-augmented control framework for legged robot locomotion, which integrates ILC with a torque library. The framework leverages the repetitive nature of periodic tasks to improve trajectory tracking while enabling rapid adaptation to nonperiodic tasks and diverse environmental conditions.

\subsection{Overview of the Hybrid Control Framework}

\textbf{The goal of the proposed control framework} is twofold: (i) to generate physically consistent motions using a trajectory optimization process grounded in full-body dynamics, and (ii) to enhance real-time tracking performance by compensating for unmodeled dynamics and disturbances through a data-enabled correction scheme.

This hybrid approach combines physics-based motion planning with ILC. Trajectory optimization provides nominal reference trajectories derived from the robot’s full-body dynamics. ILC incrementally refines the feedforward control inputs based on tracking errors observed during execution. Once convergence is achieved, the learned torques are stored in a TL and reused for rapid adaptation to similar tasks.
The framework has been implemented and validated on both the bipedal robot Cassie and the quadrupedal robot A1, demonstrating accurate and adaptable locomotion across a wide range of periodic and nonperiodic tasks.

\subsection{Hybrid System Model}

Legged locomotion is fundamentally hybrid, consisting of continuous-time dynamics punctuated by discrete transitions at contact events. These transitions occur during phase changes such as foot touchdown and lift-off, resulting in mode-dependent dynamics that must be explicitly modeled for accurate control.
To represent this behavior, we formulate the robot's motion as a hybrid dynamical system:
\begin{equation}
\label{eq:hybrid_model}
\dot{\mathbf{x}} = f(\mathbf{x}, \mathbf{u}), \quad \mathbf{x}^+ = \Delta(\mathbf{x}^-),
\end{equation}
where \( \mathbf{x} \in \mathbb{R}^n \) denotes the continuous state vector (e.g., joint positions and velocities), \( \mathbf{u} \in \mathbb{R}^m \) is the control input, \( f(\mathbf{x}, \mathbf{u}) \) defines the continuous dynamics within a contact mode, and \( \Delta(\mathbf{x}^-) \) captures the instantaneous state reset at discrete events, such as foot impacts.
This modeling framework supports both trajectory optimization and learning. During optimization, the hybrid structure is used to generate feasible and dynamically consistent motion plans. During execution, it enables segmentation of learning updates by phase, allowing for phase-specific refinement using ILC.

%
\begin{figure}[tbp]
\centering
\includegraphics[width=0.5\columnwidth]{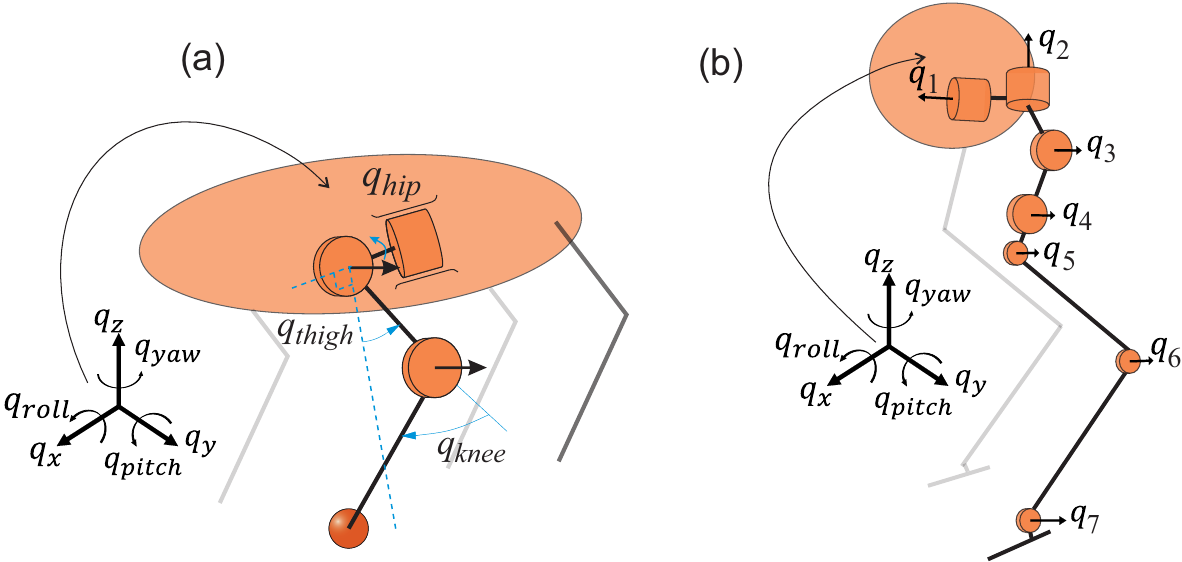}
\caption[Robot configurations]{The kinematic configurations of (a) the quadrupedal A1 \cite{alqaham2024energy} and (b) the bipedal Cassie \cite{gong2019feedback} platforms used in this study.}
\label{fig:Dynamic_Model}
\vspace{-2mm}
\end{figure}

\subsubsection{Continuous Dynamics: Stance and Swing Phases}

Legged robots such as the quadrupedal A1 and the bipedal Cassie (Figure~\ref{fig:Dynamic_Model}) exhibit hybrid dynamics characterized by alternating stance and swing phases. During the stance phase, the robot's feet are in contact with the ground, subject to holonomic constraints, while in the swing phase, the feet move freely without contact. These alternating contact conditions lead to discrete transitions between continuous dynamic states, requiring the locomotion system to be modeled as a hybrid dynamical system.

The generalized coordinates for both robot platforms are defined as:
\[
\mathbf{q} =
\begin{bmatrix}
\mathbf{q}_\mathrm{B} \\
\mathbf{q}_\mathrm{L}
\end{bmatrix}
\in \mathbb{R}^{n_\mathrm{B} + n_\mathrm{L}}.
\]
where \( \mathbf{q}_\mathrm{B} = [q_x, q_y, q_z, q_{\text{yaw}}, q_{\text{pitch}}, q_{\text{roll}}]^\mathrm{T} \in \mathbb{R}^{n_\mathrm{B}} \) denotes the base (torso) position and orientation in 3D space, and \( \mathbf{q}_\mathrm{L} \in \mathbb{R}^{n_\mathrm{L}} \) contains the actuated and passive joint angles.
For the quadrupedal A1 robot as shown in Figure~\ref{fig:Dynamic_Model}(a), each of the four legs has three actuated revolute joints: the hip, thigh, and calf. This results in \( n_\mathrm{L} = 12 \) actuated joints. For the bipedal Cassie robot in Figure~\ref{fig:Dynamic_Model}(b), each leg includes five actuated joints: a gimbal hip joint with three degrees of freedom (DOF), corresponding to hip roll(\( q_1 \)), hip yaw(\( q_2 \)),  and hip pitch(\( q_3 \)), a knee joint (\( q_4 \)), and a toe joint (\( q_7 \)). In addition, Cassie has two passive spring-loaded joints (\( q_5, q_6 \)), leading to a total of \( n_\mathrm{L} = 14 \), of which 10 are actuated.

The continuous dynamics of the robot during stance and swing phases are governed by Lagrange’s equations of motion:
\begin{equation}
\label{eq:EOM}
\operatorname{\mathbf{M}}(\mathbf{q})\ddot{\mathbf{q}} + \operatorname{\mathbf{H}}(\mathbf{q}, \dot{\mathbf{q}})\dot{\mathbf{q}} + \operatorname{\mathbf{G}}(\mathbf{q}) = \mathbf{S}\boldsymbol{\tau} + \mathbf{J}_c^\mathrm{T}\mathbf{\lambda},
\end{equation}
where:
\begin{itemize}
    \item \( \operatorname{\mathbf{M}}(\mathbf{q}) \) is the mass matrix,
    \item \( \operatorname{\mathbf{H}}(\mathbf{q}, \dot{\mathbf{q}}) \) represents Coriolis and centrifugal forces,
    \item \( \operatorname{\mathbf{G}}(\mathbf{q}) \) is the gravity vector,
    \item \( \boldsymbol{\tau} \in \mathbb{R}^{n_\mathrm{L}} \) is the joint torque vector,
    \item \( \mathbf{S} \in \mathbb{R}^{(n_\mathrm{B}+n_\mathrm{L}) \times n_\mathrm{L}} \) is the joint selection matrix,
    \item \( \mathbf{\lambda} \in \mathbb{R}^{m} \) is the ground contact wrenches, and
    \item \( \mathbf{J}_c \in \mathbb{R}^{m \times (n_\mathrm{B}+n_\mathrm{L}) } \) is the Jacobian of the holonomic contact constraints.
\end{itemize}

During the stance phase, holonomic constraints enforce that the stance foot remains stationary with respect to the ground. The constraint is written as:
\begin{equation}
\label{eq:contact_constraint}
\ddot{\mathbf{c}} = \mathbf{J}_c \ddot{\mathbf{q}} + \mathbf{\sigma} = 0,
\end{equation}
where \( \mathbf{\sigma} = \dot{\mathbf{J}}_c \dot{\mathbf{q}} \) accounts for the velocity-dependent bias acceleration.

Combining \eqref{eq:EOM} and \eqref{eq:contact_constraint}, the stance-phase dynamics are described as a differential-algebraic equation (DAE):
\begin{equation}
\label{eq:dae}
\begin{bmatrix}
    \operatorname{\mathbf{M}}(\mathbf{q}) & -\mathbf{J}_c^\mathrm{T} \\
    \mathbf{J}_c &  \mathbf{0}
\end{bmatrix}
\begin{bmatrix}
    \ddot{\mathbf{q}} \\
    \mathbf{\lambda}
\end{bmatrix}
=
\begin{bmatrix}
    \mathbf{S}\boldsymbol{\tau} - \operatorname{\mathbf{H}}(\mathbf{q}, \dot{\mathbf{q}})\dot{\mathbf{q}} - \operatorname{\mathbf{G}}(\mathbf{q}) \\
    -\mathbf{\sigma}
\end{bmatrix}.
\end{equation}

This DAE defines the function \( f(\mathbf{x}, \mathbf{u}) \) in the hybrid system model described in \eqref{eq:hybrid_model}, capturing the continuous dynamics during stance. During the swing phase, where no contact is present, the dynamics are governed purely by \eqref{eq:EOM} with \( \mathbf{\lambda} = \mathbf{0} \) and no constraint equations \cite{alqaham2024energy}.

\subsubsection{Discrete Dynamics: Touch-Down and Lift-Off}

Discrete transitions such as touch-down and lift-off introduce instantaneous changes to the system dynamics in legged locomotion. These events are modeled under the assumption of non-slipping, non-sliding, and perfectly inelastic contact. Properly handling these transitions is essential for ensuring physical fidelity and accurate control.

At touch-down, the generalized coordinates \( \mathbf{q} \) remain continuous, while the joint velocities \( \dot{\mathbf{q}} \) may change instantaneously due to the impulsive nature of contact. The hybrid state undergoes a reset:
\begin{equation}
\mathbf{x}^+ = \Delta(\mathbf{x}^-) =
\begin{bmatrix}
\mathbf{q}^- \\
\dot{\mathbf{q}}^+
\end{bmatrix},
\end{equation}
where
\[
\mathbf{x}^- =
\begin{bmatrix}
\mathbf{q}^- \\
\dot{\mathbf{q}}^-
\end{bmatrix},
\quad
\mathbf{x}^+ =
\begin{bmatrix}
\mathbf{q}^+ \\
\dot{\mathbf{q}}^+
\end{bmatrix}.
\]

The post-impact velocity \( \dot{\mathbf{q}}^+ \) is computed by enforcing holonomic contact constraints at the new contact point using the impulse-momentum equation:
\begin{equation}
\label{eq:impact_law}
\operatorname{\mathbf{M}}(\mathbf{q}) \left( \dot{\mathbf{q}}^+ - \dot{\mathbf{q}}^- \right) = \mathbf{J}_c^\mathrm{T} \mathbf{\Lambda},
\end{equation}
where \( \mathbf{\Lambda} \) is the impulse force applied during the contact event, and \( \mathbf{J}_c \) is the Jacobian of the contact constraints. Solving \eqref{eq:impact_law} ensures that the post-impact velocity satisfies \( \mathbf{J}_c \dot{\mathbf{q}}^+ = \mathbf{0} \), meaning the contacting foot has zero velocity immediately after impact. A full derivation is provided in \cite{westervelt2007feedback}.

At lift-off, the contact forces at the foot become zero and the foot enters swing. Since there is no impact, both position and velocity remain continuous:
\begin{equation}
\dot{\mathbf{q}}^+ = \dot{\mathbf{q}}^-, \quad \mathbf{\lambda} = \mathbf{0}.
\end{equation}

These discrete dynamics complete the hybrid model by coupling stance and swing phases, and are critical to replicating the physical transitions observed during legged locomotion.

\subsection{Trajectory Generation}
\label{sec:TrajGeneration}

The proposed control framework generates reference trajectories by solving a hybrid trajectory optimization problem using the open-source Fast Robot Optimization and Simulation Toolkit (FROST) \cite{hereid2017frost}. FROST formulates the problem using direct collocation with Hermite-Simpson integration, ensuring both numerical accuracy and computational efficiency. The framework supports a wide range of behaviors, including both periodic and nonperiodic tasks, across various gaits and environments.

Periodic motions, such as pronking, involve cyclic trajectories with alternating stance and flight phases. Pronking is used to evaluate tracking accuracy under symmetric, dynamic gaits. nonperiodic tasks, such as jumping or gait transitions, involve transient behaviors that demand high tracking precision, particularly for maneuvers like long jumps with tight landing margins. In these settings, feedback-driven refinement such as ILC plays a critical role in ensuring reliable execution.

\paragraph{Trajectory Optimization Formulation.}
Using FROST, we formulate the hybrid trajectory optimization as a nonlinear programming problem to generate both periodic (e.g., pronking) and nonperiodic (e.g., long jump) motions. The cost function is defined to balance energy efficiency and motion smoothness:
\begin{equation}
\label{eq:cost_function}
C = \int_{t_0}^{t_f} (\boldsymbol{\tau}^\mathrm{T} W_{\tau} \boldsymbol{\tau} + \dot{\mathbf{q}}_L^\mathrm{T} W_q \dot{\mathbf{q}}_L ) \, dt,
\end{equation}
where \( W_{\tau} \) and \( W_q \) are positive definite weighting matrices, and \( \dot{\mathbf{q}}_L \) denotes the joint velocities of the actuated joints.

The optimization is subject to the following constraints:

\vspace{2mm}
\begin{tcolorbox}[colback=gray!5!white, colframe=gray!80!black, title=General Constraints]
\begin{itemize}
    \item Dynamic consistency: Enforced through the system dynamics in~\eqref{eq:EOM};
    \item Holonomic constraints: Ensure fixed foot position during stance, as in~\eqref{eq:contact_constraint};
    \item Unilateral contact: \( \mathbf{g}_l(\mathbf{q}) > 0 \) to keep feet above ground during flight;
    \item Joint configuration limits: \( \mathbf{q}_{\min} \leq \mathbf{q} \leq \mathbf{q}_{\max} \);
    \item Joint velocity limits: \( |\dot{\mathbf{q}}| \leq \dot{\mathbf{q}}_{\max} \);
    \item Actuation limits: \( |\boldsymbol{\tau}| \leq \boldsymbol{\tau}_{\max} \);
    \item Friction cone: \( \|\mathbf{\lambda}_t\|_2 \leq \mu \|\mathbf{\lambda}_n\|_2 \), where \( \mathbf{\lambda}_t \) and \( \mathbf{\lambda}_n \) denote the tangential and normal components of the contact force, respectively, and \( \mu \) is the coefficient of friction.
\end{itemize}
\end{tcolorbox}

\begin{tcolorbox}[colback=gray!5!white, colframe=gray!80!black, title=Task-Specific Constraints]
\textbf{Periodic Motion (Pronking Gait):}
\begin{itemize}
    \item Average forward speed: \( \bar{q}_{x} = \frac{q_{x}(t_f) - q_{x}(t_0)}{t_f - t_0} \);
    \item Maximum torso height: \( q_{z} \leq 0.34 \, \mathrm{m} \);
    \item Gait periodicity: \( \mathbf{q}(t_0) = \mathbf{q}(t_f) \), \( q_x(t_0) \neq q_x(t_f) \), where \( t_0 \) and \( t_f \) denote the start and end times of the gait cycle, respectively. 
\end{itemize}

\textbf{nonperiodic Motion (Long Jump):}
\begin{itemize}
    \item Horizontal jumping distance: \( d = 0.35 \, \mathrm{m} \);
    \item Initial configuration: \( \mathbf{q}(t_0) = \mathbf{q}_i \);
    \item Final configuration: \( \mathbf{q}(t_f) = \mathbf{q}_f \);
    \item Resting condition at landing: \( \dot{\mathbf{q}}(t_f) = \mathbf{0} \).
\end{itemize}
\end{tcolorbox}

\subsubsection{Bezier Polynomial Representation}
The joint trajectories are approximated using Bézier polynomials to ensure smooth transitions and compact representation:
\begin{equation}
\label{eq:bezier_polynomials}
   h_{j}(s) = \sum_{i=0}^{n_M} \alpha_{j,i} \frac{n_M !}{i !(n_M-i) !} s^{i}(1-s)^{n_M-i},
\end{equation}
where \( s \in [0,1] \) is the phase variable, \( \alpha_{j,i} \) are Bézier coefficients, and \( n_M \) is the polynomial order. The phase variable \( s \) transitions from 0 (start of phase) to 1 (end of phase). Each optimal solution is characterized by a matrix \( \mathbf{B} \in \mathbb{R}^{n_L \times (n_M+1)} \), containing the coefficients for all joints. This matrix can be parametrized by specific tasks such as the running velocities \( \bar{q}_{x} \) or jumping distance \( d \), and in the following parts we refer to it as a set of \textbf{motion primitives}.

\subsection{Controller Structure}

\begin{figure*}[tbp]
\centering
\includegraphics[width=1\columnwidth]{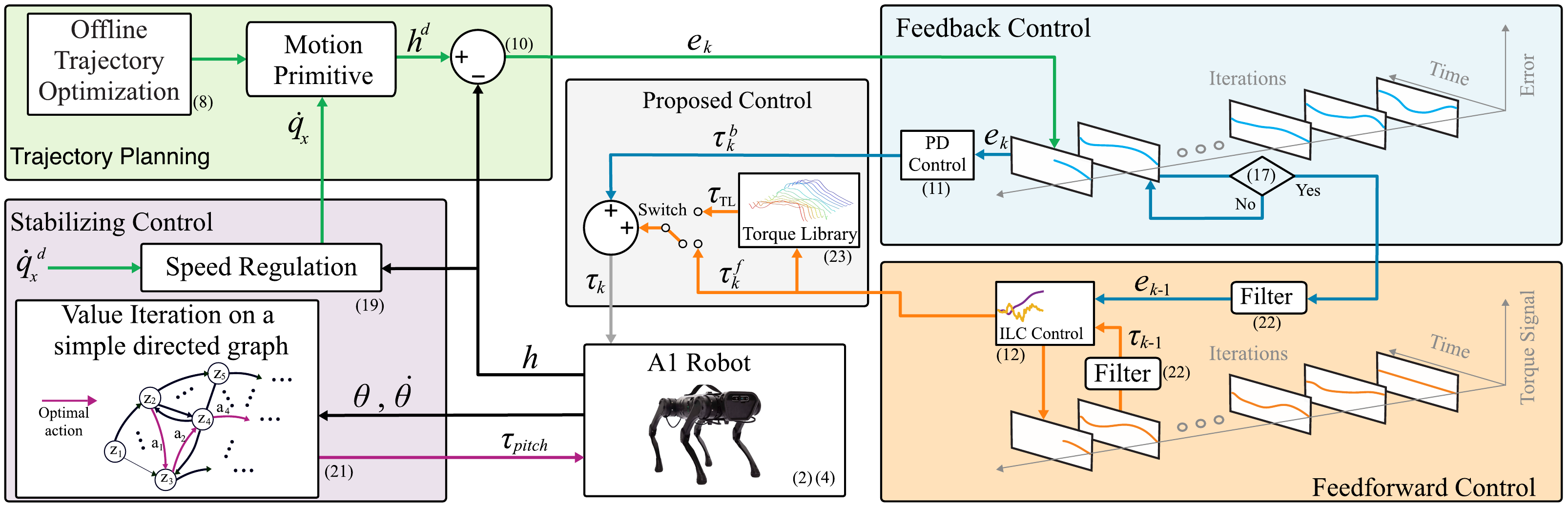}
\caption[Hierarchical controller structure]{The proposed control architecture includes trajectory planning (green), feedback control (blue), and feedforward control (orange) modules. An iterative policy further improves stability by refining torques applied to the thigh joints. Zero-phase filtering ensures smooth and phase-consistent control signals. Key computations align with the equations presented in this section. 
}
\label{fig:controller_structure}
\vspace{-2mm}
\end{figure*}

Optimized solutions are stored in a motion primitive library and retrieved during online execution to serve as reference trajectories. For periodic motions, trajectories are replayed in synchronization with predefined gait phases. For nonperiodic tasks, stored solutions are used as initializations and refined adaptively during execution.
The controller follows a hierarchical structure integrating trajectory planning, feedback control, and feedforward compensation, as shown in Figure~\ref{fig:controller_structure}. Each module is visually distinguished to emphasize its function and interaction with the others. In addition to these core components, a learned iterative policy is introduced to further enhance torso stabilization, with corrections applied primarily at the thigh joints. A zero-phase filter is used to smooth the resulting torque profiles, eliminating phase delays and unrealistic components.
Key computations referenced in this section are annotated within the figure for clarity.

\subsubsection{Trajectory Tracking with ILC}

The proposed trajectory tracking controller integrates two complementary components to achieve accurate and adaptive locomotion. The first component is a feedback controller, implemented as a proportional-derivative (PD) regulator, which addresses tracking errors in real time. This controller provides immediate stabilization and disturbance rejection, ensuring robustness during execution despite model inaccuracies or external perturbations.
The second component is a feedforward controller based on ILC. ILC updates the feedforward torque profiles by leveraging tracking error data from previous executions. This allows the system to learn from past mistakes and compensate for unmodeled dynamics or transient disturbances. Over successive iterations, ILC enables rapid improvement in tracking accuracy with minimal data requirements. As illustrated in Figure~\ref{fig:controller_structure}, the feedback and feedforward modules operate concurrently, with their interactions forming the foundation of the control hierarchy.
These two control components—highlighted in blue (feedback) and orange (feedforward) in Figure~\ref{fig:controller_structure}—operate in parallel to deliver precise and adaptive control. 

\paragraph{Mathematical Formulation of Trajectory Tracking and Feedback Control:} 
Let the desired joint-space trajectories be represented by \( \mathbf{h}_{\text{d}}(\mathbf{B}, s) \in \mathbb{R}^{n_{L}} \), where \( \mathbf{B} \in \mathbb{R}^{n_L \times (n_M + 1)} \) contains the Bézier coefficients for all joints. 
The trajectory tracking problem at the \( k \)-th iteration is defined using \( n_L \) virtual constraints, which encode the difference between the desired and actual joint trajectories. The tracking error \( \mathbf{e}_k(s) \) is expressed as:
\begin{equation}
\label{eq:virtualconstraint}
   \mathbf{e}_k(s) = \mathbf{h}^{\text{d}}(\mathbf{B}, s) - \mathbf{h}(\mathbf{q}_{L}, t_k/T),
\end{equation}
where \( t_k \) is the elapsed time within the current phase and \( T \) is the task duration. These virtual constraints guide the system toward reducing the joint tracking error throughout the phase.
The feedback control torque \( \boldsymbol{\tau}^{b}_{k}(s) \) compensates for deviations in real-time and is computed as:
\begin{equation}
\label{eq:feedback}
   \boldsymbol{\tau}^{b}_{k}(s) = K_P^{b} \mathbf{e}_k(s) + K_D^{b} \dot{\mathbf{e}}_k(s),
\end{equation}
where \( K_P^{b} \) and \( K_D^{b} \) are the proportional and derivative gain matrices, respectively, and remain fixed across all iterations. This feedback controller enhances the system's robustness to sensor noise, modeling error, and transient disturbances during motion execution.

\paragraph{Feedforward Control via ILC:}
ILC leverages the repetitive structure of tasks such as pronking and trotting to iteratively improve feedforward control. By utilizing tracking errors from previous executions, it adjusts control inputs to better account for unmodeled dynamics and external disturbances. The feedforward torque \( \boldsymbol{\tau}^{f}_{k}(s) \) at iteration \( k \) is computed as:
\begin{align}
\label{eq:feedforward}
   \boldsymbol{\tau}^{f}_{k}(s) = \boldsymbol{\tau}_{k-1}(s) &+ K_P^{f} \mathbf{e}_{k-1}(s + \delta s) \\ \notag
   &+ K_D^{f} \dot{\mathbf{e}}_{k-1}(s + \delta s),
\end{align}
where \( \boldsymbol{\tau}_{k-1}(s) \) is the applied torque from the previous iteration, \( K_P^{f} \) and \( K_D^{f} \) are task-specific proportional and derivative gain matrices, and \( \delta s \) introduces a phase lead to anticipate and pre-compensate future tracking errors.
The total control input applied to the robot is the sum of feedback and feedforward components:
\begin{equation}
\label{eq:desiredtorque}
   \boldsymbol{\tau}_{k}(s) = \boldsymbol{\tau}^{b}_{k}(s) + \boldsymbol{\tau}^{f}_{k}(s),
\end{equation}
which defines the joint torques applied during the \( k \)-th execution. Initially, only feedback control is applied based on the reference trajectory. Once ILC is activated, the feedforward term is initialized from the previous feedback torque as \( \boldsymbol{\tau}^{f}_{1}(s) = \boldsymbol{\tau}^{b}_{0}(s) \), and \eqref{eq:desiredtorque} is used to iteratively refine performance over repeated executions.

\paragraph{Comparison with WBC:}
In WBC frameworks \cite{Katz_2019}, feedforward torques are computed using inverse dynamics:
\begin{equation}
\label{eq:WBC}
   \boldsymbol{\tau} 
   = \mathbf{S}^\mathrm{T} \left( \operatorname{\mathbf{M}} \ddot{\mathbf{q}}^{d} + \operatorname{\mathbf{H}} \dot{\mathbf{q}} + \operatorname{\mathbf{G}} - \mathbf{J}_c^\mathrm{T} \mathbf{\lambda}^{d} \right),
\end{equation}
where \( \ddot{\mathbf{q}}^{d} \) is the desired joint acceleration, and \( \mathbf{\lambda}^{d} \) is the desired GRF. This method provides accurate torque computation when model parameters are well-calibrated. However, its performance deteriorates when there are discrepancies between the assumed and actual dynamics, such as unmodeled friction, terrain compliance, or joint backlash.
In contrast, the proposed ILC controller learns the corrective feedforward torques directly from tracking errors observed over repeated executions. By refining the torque profiles iteratively, ILC achieves high tracking accuracy without relying on detailed physical models. Moreover, once the motion is learned, ILC enables extremely fast execution by eliminating the need for online model inversion.
Compared to WBC, the ILC-based controller offers several advantages:
\begin{itemize}
    \item \textbf{Accuracy:} Both methods can achieve high tracking precision, but ILC improves performance over time through data-driven refinement, especially in the presence of modeling errors.
    \item \textbf{Adaptability:} ILC readily adapts to variations in mass properties, contact conditions, and unmodeled internal dynamics, whereas WBC performance tends to degrade in the presence of model inaccuracies.
    \item \textbf{Computational Efficiency:} Once learned, ILC executes with negligible runtime cost, whereas WBC requires solving inverse dynamics with contact constraints at every control step.
\end{itemize}

\paragraph{Performance Metrics:}
To evaluate the tracking accuracy of the controller, we compute the root-mean-square error (RMSE) for each joint over a trajectory sampled at \( N_s \) uniformly spaced points along the phase variable \( s \in [0, 1] \). The RMSE for joint \( \ell \) at iteration \( k \) is defined as:
\begin{equation}
\label{eq:rmse_joint}
\mathrm{RMSE}_{\ell,k} = \sqrt{\frac{1}{N_s} \sum_{i=1}^{N_s} \left(h_{\ell,k}(s_i) - h^{d}_{\ell,k}(s_i)\right)^2},
\end{equation}
where \( h_{\ell,k}(s_i) \) and \( h^{d}_{\ell,k}(s_i) \) denote the actual and desired joint trajectories evaluated at the phase sample \( s_i \). The average tracking error across all \( n_L \) actuated joints is then given by:
\begin{equation}
\label{eq:rmse_avg}
\Delta_k = \frac{1}{n_L} \sum_{\ell=1}^{n_L} \mathrm{RMSE}_{\ell,k}.
\end{equation}

We use this metric to assess whether an iteration improves performance. A new trial is accepted for learning only if:
\begin{equation}
\label{eq:improve_metric}
\Delta_k < \Delta_0 + \left( \delta_e - \Delta_0 \right) \cdot \frac{2}{\pi} \cdot \tan^{-1}(\mu_e k),
\end{equation}
where \( \delta_e \) is the terminal RMSE tolerance and \( \mu_e \) is a positive shaping parameter.
Finally, the learning process is terminated once the improvement criterion is satisfied in \( N_e \) iterations:
\begin{equation}
\label{eq:stop_condition}
\sum_{k=3}^{K} \boldsymbol{1}\left( \Delta_k < \kappa \delta_e \right) = N_e,
\end{equation}
where \( \kappa \) is a tunable margin factor, and \( \mathbb{1}(\cdot) \) is the indicator function.

\subsubsection{Stabilizing Controller Design}
\paragraph{Speed Regulation.}
Stabilization of longitudinal motion is achieved using a speed regulation strategy inspired by Raibert's foot-placement control approach \citep[Chapter~3]{raibert1986legged}. This controller dynamically adjusts the leg touchdown angle in response to velocity deviations, helping maintain the desired average forward speed.
The controller computes a corrected desired leg angle \( \theta^d \)\footnote{For the A1 robot, the leg angle in the sagittal plane is defined as:
\[
\theta = q_{\text{thigh}} + \frac{1}{2} q_{\text{calf}}, 
\]
where \( q_{\text{thigh}} \) and \( q_{\text{calf}} \) are the respective joint angles. Similar kinematic formulations for the Cassie robot are detailed in \citep[(15)]{gong2019feedback}, including the mapping from desired leg motion to individual joint trajectories.} just before ground contact using:
\begin{equation}
\label{eq:FPC}
\theta^{d} = \theta + K_{\theta} \left( \dot{q}_x - \dot{q}_x^{d} \right),
\end{equation}
where \( \dot{q}_x \) is the current torso velocity in the sagittal direction, \( \dot{q}_x^{d} \) is the desired velocity, and \( K_{\theta} \) is a proportional gain. This feedback mechanism compensates for disturbances and corrects velocity drift by adjusting foot placement.
The updated leg angle \( \theta_d \) is then mapped back to joint space using inverse kinematics. Reference trajectories for all joints are subsequently updated from the motion primitive library based on the corrected velocity \( \dot{q}_x \). 
Further details on related feedback-interpolation schemes can also be found in \cite{da20162d, gong2019feedback}.

\paragraph{Attitude Control.}
\label{sec:attControl}
Maintaining a stable torso orientation—particularly in pitch—is critical for dynamic legged locomotion. This task becomes increasingly challenging during high-speed gaits, where stance durations are brief (e.g., approximately 0.2 seconds) and ground contact conditions may vary across legs. To complement the ILC-based trajectory tracking strategy, we implement a value iteration-based controller that stabilizes torso attitude through optimal feedback.
The robot’s reduced-order attitude state is represented by \( \mathbf{z} \in \mathbb{R}^{n_z} \), which captures relevant torso orientation components, estimated via a Kalman filter. At each time step, a corrective input \( \mathbf{a} = \pi(\mathbf{z}) \) is applied, where \( \pi(\cdot) \) is the optimal policy designed to minimize long-term deviation from a desired state \( \mathbf{z}^d \).
We collect the states (or nodes) and the inputs (or actions) in discrete-time system into \(\mathcal{Z} = \left\{ \mathbf{z}_1, \mathbf{z}_2, \mathbf{z}_3,\ldots \right\}\) and \(\mathcal{A} = \left\{ \mathbf{a}_1, \mathbf{a}_2, \mathbf{a}_3,\ldots \right\}\)

The policy is obtained by minimizing the following infinite-horizon cost function:
\begin{equation}
\label{eq:attitude_cost}
V(\mathbf{z}) = \min \sum_{n_t=0}^{\infty} \left[ (\mathbf{z}_{n_t} - \mathbf{z}^{d})^\mathrm{T} Q (\mathbf{z}_{n_t} - \mathbf{z}^{d}) + \gamma \mathbf{a}_{n_t}^2 \right],
\end{equation}
where \( Q \in \mathbb{R}^{n_z \times n_z} \) is a positive-definite matrix penalizing deviation from the desired attitude, and \( \gamma > 0 \) is a scalar that penalizes control effort.
The optimal control policy is computed offline using value iteration over a discretized domain of the system dynamics \( g(\mathbf{z}, \mathbf{a}) \) and stage cost \( r(\mathbf{z}, \mathbf{a}) \). The feedback control law is given by:
\begin{equation}
\label{eq:optimal_attitude_policy}
\pi^*(\mathbf{z}_{n_t}) = \arg\min_{\mathbf{a} \in \mathcal{A}} \left[ r(\mathbf{z}_{n_t}, \mathbf{a}) + V^*\left(g(\mathbf{z}_{n_t}, \mathbf{a})\right) \right],
\end{equation}
where \( V^*(\mathbf{z}) \) denotes the optimal value function over the discretized state space.
This attitude regulation scheme ensures the pitching motion remains bounded despite terrain disturbances or asymmetrical ground contact. A detailed derivation of value iteration for robotic control can be found in \citep[Chapter~6]{tedrakeUnderactuated}. 

\subsection{Zero-Phase Filter Implementation}
\label{sec:filter}

Minimizing phase delays is critical in ILC for precise trajectory tracking, especially in legged robotics where rapid dynamics demand accurate control inputs. To address this, our framework employs an offline zero-phase filtering strategy applied after each task execution. This approach eliminates phase distortion, preserving signal integrity for the iterative learning process.

Zero-phase filtering is achieved by processing the data bidirectionally using a filter. In our implementation, we utilize a forward-backward Infinite Impulse Response (IIR) filter, defined by its discrete transfer function \(H(z)\). This bidirectional processing effectively cancels out the phase response of the filter. If \(X(e^{j\omega})\) is the input signal and \(H(e^{j\omega})\) is the filter's frequency response, the zero-phase filtered output \(Y(e^{j\omega})\) in the frequency domain can be represented as:

\begin{equation}
\label{eq:filtfilt_out_concise}
Y\left( e ^{j \omega}\right) = X\left( e ^{ j \omega}\right)\left|H\left( e ^{ j \omega}\right)\right|^2.
\end{equation}

This ensures that the output signal is temporally aligned with the input, removing phase shifts while retaining the desired frequency characteristics.

The offline application of zero-phase filtering, performed after each stride for periodic motions or after each task for nonperiodic motions, ensures that the signals used for ILC updates are free from phase delays without introducing real-time computational overhead. For periodic gaits, filtered data from the current segment informs the next iteration. For nonperiodic tasks, filtered data is stored for future consistent performance.
By providing temporally aligned filtered torque \( \boldsymbol{\tau}_k(s) \) and error \( \mathbf{e}_k(s) \) signals, zero-phase filtering enhances the stability and convergence of our ILC controller, leading to more accurate feedforward updates and reduced iteration-to-iteration variability. This is particularly beneficial for high-precision tasks like dynamic gait transitions, where timing accuracy is paramount. Complementing this, we recalculate the robot's torso velocity during stance using leg odometry \cite{6094484} (assuming no slip) to further mitigate sensor noise. This offline zero-phase filtering strategy is a practical solution for maintaining signal fidelity in both periodic and nonperiodic robotic tasks within the iterative control framework.

\subsection{Integration with Torque Library}
\label{sec:TorqueLibrary}

The TL serves as a repository for optimized feedforward torque profiles generated by the ILC framework. As described in \eqref{eq:desiredtorque}, torques are refined iteratively using tracking errors from previous executions. Once convergence is achieved—quantified by the stopping condition in~\eqref{eq:stop_condition}—the resulting torque trajectory is stored in the TL to support future reuse.

Each stored profile corresponds to a specific locomotion task (e.g., gait type, speed, terrain) and is parameterized by a matrix of Bézier coefficients, as defined in~\eqref{eq:bezier_polynomials}. This representation enables compact storage and smooth interpolation across similar tasks. Alongside each profile, metadata including average speed, joint trajectories, and contact timing is recorded to facilitate efficient retrieval.

During online execution, the TL is queried using current task parameters. If a matching entry is found, the associated torque coefficients are directly used as the initial feedforward command in~\eqref{eq:desiredtorque}. When the task falls between stored conditions, torque initialization is performed by linear interpolation between neighboring entries:
\begin{equation}
\label{eq:TL_interp}
\mathbf{T}_{\tau} = \frac{p_b - p}{p_b - p_a} \mathbf{T}_{\tau}^{(a)} + \frac{p - p_a}{p_b - p_a} \mathbf{T}_{\tau}^{(b)},
\end{equation}
where \( p \) denotes the current task parameter (e.g., desired speed), and \( \mathbf{T}_{\tau}^{(a)} \), \( \mathbf{T}_{\tau}^{(b)} \) are Bézier coefficient matrices for the closest stored profiles with \( p_a \leq p < p_b \). This approach mirrors the interpolation used for trajectory generation in speed regulation similar to \citep[(8)\&(9)]{da20162d}, ensuring smooth transitions in feedforward control.

The TL supports both periodic and nonperiodic tasks. For periodic gaits (e.g., pronking, trotting), torque profiles are aligned with gait phase \( s \in [0,1] \), allowing consistent application over repeated strides. For nonperiodic maneuvers (e.g., jumping, recovery), stored torques act as high-quality initializations that accelerate convergence during ILC refinement.
As the robot encounters new environments or tasks, the TL evolves by incorporating newly learned profiles. This incremental learning process mirrors biological motor adaptation and enables long-term generalization. The modular design of the TL also supports scalability across platforms, allowing deployment on robots with different morphologies.

Overall, the integration of the TL into the control framework provides three key benefits:
\begin{itemize}
    \item \textbf{Reduced online computation}, by reusing previously optimized torque profiles and eliminating redundant learning;
    \item \textbf{Accelerated task adaptation}, through interpolation-based initialization of feedforward control;
    \item \textbf{Improved scalability}, by supporting generalization across different speeds, gaits, and robotic platforms.
\end{itemize}

These capabilities establish the TL as a critical component of the proposed control architecture, particularly in enabling efficient and adaptive feedforward strategies for both periodic and nonperiodic locomotion tasks.

\section{Simulation and Experimental Results}
\label{sec:results}

This section presents simulation and experimental validation of the proposed control framework on both the quadrupedal A1 and the bipedal Cassie robots across a range of periodic and nonperiodic tasks. Simulations for A1 were conducted in the Gazebo environment using ROS \cite{koenig2004design}, which provides high-fidelity modeling of robot dynamics. To assess robustness, environmental parameters such as terrain slope and gravitational constant were systematically varied.

Experimental tests on A1 were performed across diverse terrain types, including indoor carpet, outdoor concrete, grass, snow-covered pavement, and public roads with varying inclines. Additionally, a range of walking speeds was tested on the Cassie Blue robot at the University of Michigan. Across both platforms, the proposed framework rapidly improved motion accuracy, with convergence observed within seconds of controller activation. These results highlight the method’s adaptability and robustness in real-world conditions.
Compared to the state-of-the-art WBC approach, the proposed method achieved significantly better trajectory tracking, eliminated phase delays, and improved computational efficiency by an order of magnitude.

The following subsections provide quantitative and qualitative evaluations across varying tasks, speeds, terrains, and simulated gravitational environments. Results are drawn from both simulation and hardware, and demonstrate the generality of the proposed framework.

\subsection{Simulation Validation of Iterative Locomotion Improvements Across Varying Speeds and Gravitational Conditions}

To validate the proposed control framework, simulations were conducted using the A1 quadrupedal robot executing a pronking gait under conditions designed to replicate realistic dynamics. Environmental parameters such as ground friction and contact restitution were carefully tuned to reflect real-world behavior. These simulations offered a safe, repeatable platform for fine-tuning control parameters and evaluating the system’s adaptability.
The simulations demonstrated that iteratively refining feedforward torques through ILC significantly improved joint trajectory tracking. Initial experiments were performed on flat terrain to establish a performance baseline, followed by tests under varying gravitational fields and uneven surfaces. The controller operated at 1~kHz, and all computations were executed in real time on a high-performance desktop computer equipped with an Intel\textregistered{} Core\textsuperscript{TM} i7 processor.

\begin{figure}[tbp]
\centering
\includegraphics[width=0.5\columnwidth]{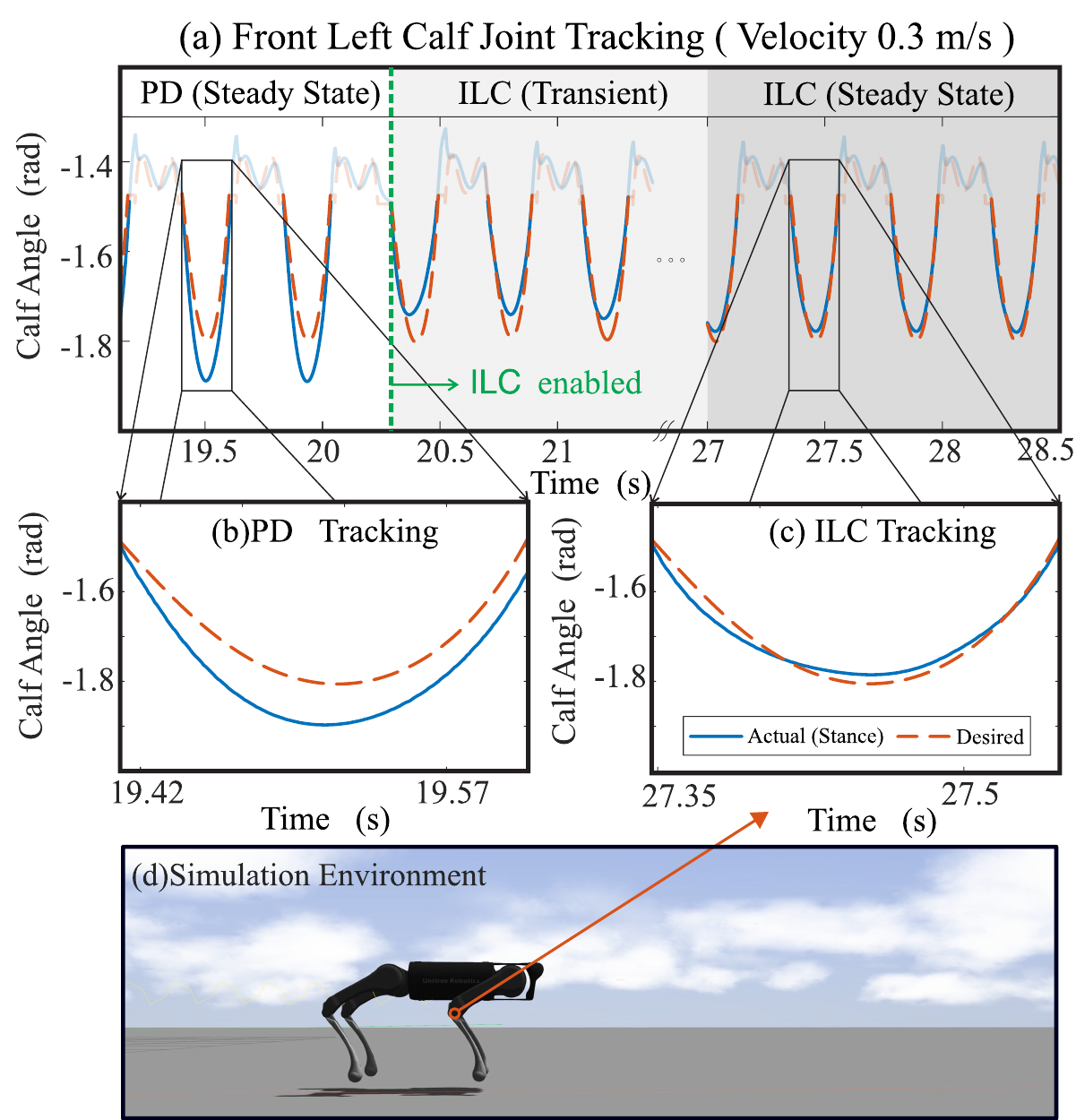}
\caption{Simulated tracking results for the front left calf joint of A1 during pronking. When ILC was activated at \( t = 20.3 \, \si{\second} \), the joint tracking error decreased by 83.6\% within 7 seconds, demonstrating rapid convergence compared to baseline PD control. (d) Gazebo, a physics-based simulation environment.} 
\label{fig:ILC_SIM_Calf}
\vspace{-2mm}
\end{figure}
%

\subsubsection{Iterative Learning for Enhanced Locomotion Across Variable Speeds}

The simulation dataset includes joint angles, velocities, torques, ground reaction forces, and tracking errors across 14 distinct pronking trajectories, each designed for a different average speed. These speeds range from $-0.6$ to $0.8$~\si[per-mode=symbol]{\meter\per\second} in increments of $0.1$~\si[per-mode=symbol]{\meter\per\second}. Each trajectory is parameterized using B\'ezier polynomials to ensure smooth and differentiable motion profiles.

Initially, a PD controller was employed to track reference trajectories generated from the motion primitive library. As shown in Figure~\ref{fig:ILC_SIM_Calf}, the PD controller maintained approximate adherence but exhibited noticeable tracking errors in the calf joint, particularly during the mid-stance phase. These errors, reaching up to \SI{0.09}{\radian}, are primarily attributed to the demands of gravitational compensation and unmodeled actuator dynamics.

The results confirmed that PD control alone is insufficient for high-accuracy tracking, especially under dynamic conditions such as speed variation and terrain uncertainty. To address this limitation, the ILC algorithm was activated at \( t = 20 \)~\si{\second}, with feedforward control taking effect in the subsequent stance phase (\( t = 20.3 \)~\si{\second}). Over the next 17 strides (\( \sim 7 \)~\si{\second}), the calf joint error was reduced by 83.6\%. On average, ILC reduced tracking errors by 75.8\% in the calf joint and 30.8\% in the thigh joint across all trajectories.
These results demonstrate the adaptability of the ILC framework to varying locomotion speeds. 
\begin{figure}[tbp]
\centering
\includegraphics[width=0.5\columnwidth]{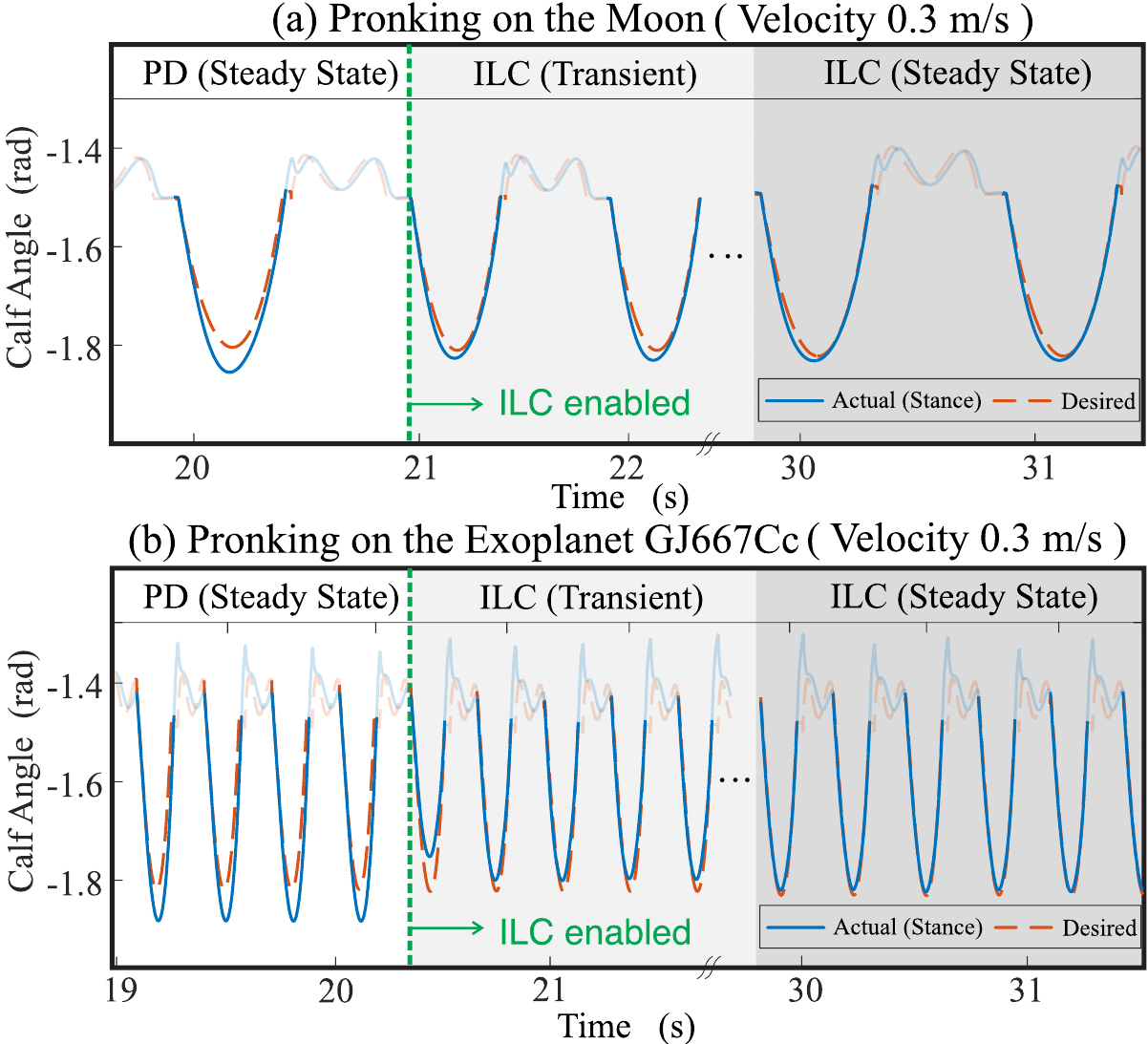}
\caption{Tracking improvements for the A1 robot's calf joints during pronking under lunar and high-gravity conditions. The implementation of ILC achieves significant error reductions in both scenarios, highlighting its adaptability and robustness across diverse gravitational environments.}
\label{fig:moon}
\vspace{-2mm}
\end{figure}

\subsubsection{Iterative Learning Under Different Gravity Conditions}
To evaluate the generalizability of the proposed control framework, simulations were conducted in environments with varying gravitational forces:
\begin{itemize}
    \item Lunar Gravity: \( 1.62 \, \si[per-mode=symbol]{\meter\per\second^2} \).
    \item Exoplanet GJ667Cc Gravity: \( 15.70 \, \si[per-mode=symbol]{\meter\per\second^2} \).
\end{itemize}
The stride time \( T_\mathrm{planet} \) was adjusted for each environment using the relationship based on the scaling effects \cite{Hof:1996}:
\[
T_\mathrm{planet} = T_\mathrm{earth} \sqrt{\frac{g_\mathrm{earth}}{g_\mathrm{planet}}},
\]
where \( g_\mathrm{earth} = 9.81 \) \si[per-mode=symbol]{\meter\per\second^2}. 
The dataset includes joint trajectories, tracking errors, and ground reaction forces for each gravitational condition. In lunar gravity, the PD controller performed better due to reduced gravitational demands, but ILC further reduced tracking errors by 34\%. In the high-gravity scenario, ILC achieved a 53\% reduction in tracking errors within 9 strides. Figure~\ref{fig:moon} highlights these improvements, demonstrating the adaptability of the controller across extreme gravitational environments.

\subsection{Hardware Testing for Learning Locomotion Across Varying Terrains}

Building upon the simulation results, the proposed ILC framework was deployed on hardware platforms, specifically, the quadrupedal A1 and the bipedal Cassie to evaluate its performance in real-world scenarios. Transitioning from simulation to hardware introduces additional challenges, including unknown friction coefficients, unmodeled actuator dynamics, and external disturbances. These factors necessitate thorough validation and fine-tuning of the control algorithm to ensure robust and stable operation.
On the A1 robot, a Raspberry Pi\textregistered{} 4 embedded in the torso processes sensor data, while an external laptop equipped with an Intel\textregistered{} Core\textsuperscript{TM} i5 processor executes control commands at a frequency of 1~kHz. This setup supports high-frequency control suitable for dynamic locomotion.

Experiments were conducted on various terrains including flat indoor surfaces, inclined slopes, and grass fields to evaluate the controller's adaptability. These environments presented a diverse set of ground contact conditions and disturbances, offering a rigorous test of the framework's adaptability. A key element of the control strategy was whole-body attitude stabilization, implemented using the value-iteration-based feedback policy described in Section~\ref{sec:attControl}. Special attention was given to pitch regulation, which plays a dominant role in maintaining balance during dynamic, nonperiodic motions such as pronking.
The torso pitch dynamics were modeled as a planar rigid body rotating about the lateral axis, with a total moment of inertia \( I_{\text{pitch}} = 0.038 \, \mathrm{kg \cdot m^2} \). The input state to the stabilizing controller consisted of the torso's pitch angle and angular velocity, represented by the reduced state vector:$\vec{z} = [ q_{\text{pitch}},\dot{q}_{\text{pitch}}]^\mathrm{T}$.
The feedback control input was then computed via the optimal policy \( a = \pi(\vec{z}) \), enabling precise regulation of torso orientation throughout the learning process.

\subsubsection{Locomotion Across Natural Terrains}

\begin{figure}[tbp]
\centering
\includegraphics[width=0.5\columnwidth]{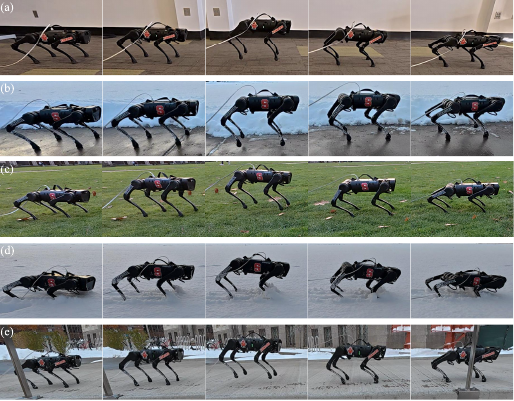}
\caption[outdoor]{A1 robot performing locomotion using the hybrid control scheme across diverse terrains: (a) indoor carpet, (b) wet outdoor concrete, (c) natural grass, (d) snow-covered surface, and (e) inclined ground.}
\label{fig:slopes}
\vspace{-2mm}
\end{figure}

To assess the robustness of the proposed control framework in real-world conditions, hardware experiments were conducted across five distinct terrain types: indoor carpet, wet outdoor concrete, natural grass, snow-covered pavement, and inclined surfaces, as shown in Figure~\ref{fig:slopes}. These terrains present diverse challenges, including surface compliance, friction variability, and unmodeled irregularities, making them ideal for validating the generalization of the learned control policy\footnote{\url{https://github.com/DLARlab/MuscleMemory_for_HighPrecisionLocomotion}}.

Across all terrain types, the controller exhibited stable performance and significant tracking improvements with ILC compared to PD control. For example, on flat grassy terrain characterized by uneven ground and hidden depressions, the joint tracking RMSE decreased from \( \SI{0.11}{\radian} \) and \( \SI{0.06}{\radian} \) for the calf and thigh joints under PD control to \( \SI{0.03}{\radian} \) and \( \SI{0.04}{\radian} \) with ILC, respectively. Similar improvements were observed on inclined surfaces and snow-covered ground, demonstrating the controller's adaptability to disturbances and unstructured environments.

\subsubsection{Learning Precise Motions for A1 and Cassie Robots on Flat Ground}

\begin{table}[tbp]
\caption{Tracking performance improvements for the A1 robot across varying speeds. The error reduction percentage quantifies the improvement achieved by ILC compared to PD control.}
\label{table:A1SpeedTracking}
\setlength{\tabcolsep}{0.35em} 
\renewcommand{\arraystretch}{1.1}
\begin{center}
\vspace{-2mm}
\begin{tabular}{|c||c|c|r||c|c|r|}
\hline 
\multirow{2}{1cm}{\centering{\textbf{Speed (m/s)}}}  & \multicolumn{3}{c||}{\textbf{Calf Error (rad)}} & \multicolumn{3}{c|}{\textbf{Thigh Error (rad)}}  \\  \cline{2-7} 
& \textit{PD}  & \textit{ILC} & \textit{Reduction} & \textit{PD} & \textit{ILC} & \textit{Reduction} \\ 
\hline
-0.6  & 0.11  &  0.03  &  72.7\%   &  0.24  &  0.06  & 75.0\%  \\
-0.5  & 0.14  &  0.02  &  85.7\%   &  0.15  &  0.07  & 53.3\%  \\  
-0.4  & 0.14  &  0.02  &  85.7\%   &  0.14  &  0.06  & 57.1\%  \\ 
-0.3  & 0.13  &  0.03  &  76.9\%   &  0.12  &  0.04  & 66.7\%  \\ 
-0.2  & 0.12  &  0.03  &  75.0\%   &  0.12  &  0.05  & 58.3\%  \\ 
-0.1  & 0.10  &  0.03  &  70.0\%   &  0.10  &  0.04  & 60.0\%  \\  
0.0   & 0.11  &  0.04  &  63.6\%   &  0.06  &  0.05  & 16.7\%  \\
0.1   & 0.11  &  0.05  &  54.5\%   &  0.07  &  0.05  & 28.6\%  \\
0.2   & 0.11  &  0.03  &  72.7\%   &  0.05  &  0.03  & 41.7\%  \\
0.3   & 0.10  &  0.03  &  70.0\%   &  0.07  &  0.04  & 42.9\%  \\
0.4   & 0.12  &  0.05  &  58.3\%   &  0.08  &  0.06  & 25.0\%  \\
0.5   & 0.11  &  0.05  &  54.5\%   &  0.06  &  0.05  & 16.7\%  \\
0.6   & 0.10  &  0.06  &  40.0\%   &  0.08  &  0.06  & 25.0\%  \\
0.7   & 0.09  &  0.06  &  33.3\%   &  0.08  &  0.06  & 25.0\%  \\
0.8   & 0.13  &  0.07  &  38.5\%   &  0.08  &  0.06  & 25.0\%  \\
\hline
\end{tabular}
\vspace{-6mm}
\end{center}
\end{table}

Table~\ref{table:A1SpeedTracking} reports the RMSE tracking improvements achieved by ILC over PD control for the A1 robot across 15 pronking speeds ranging from \SI[per-mode=symbol]{-0.6}{\meter\per\second} to \SI[per-mode=symbol]{0.8}{\meter\per\second}. Substantial reductions in RMSE are observed in both calf and thigh joints, particularly at lower speeds.

\begin{figure*}[tbp]
\centering
\includegraphics[width=1\columnwidth]{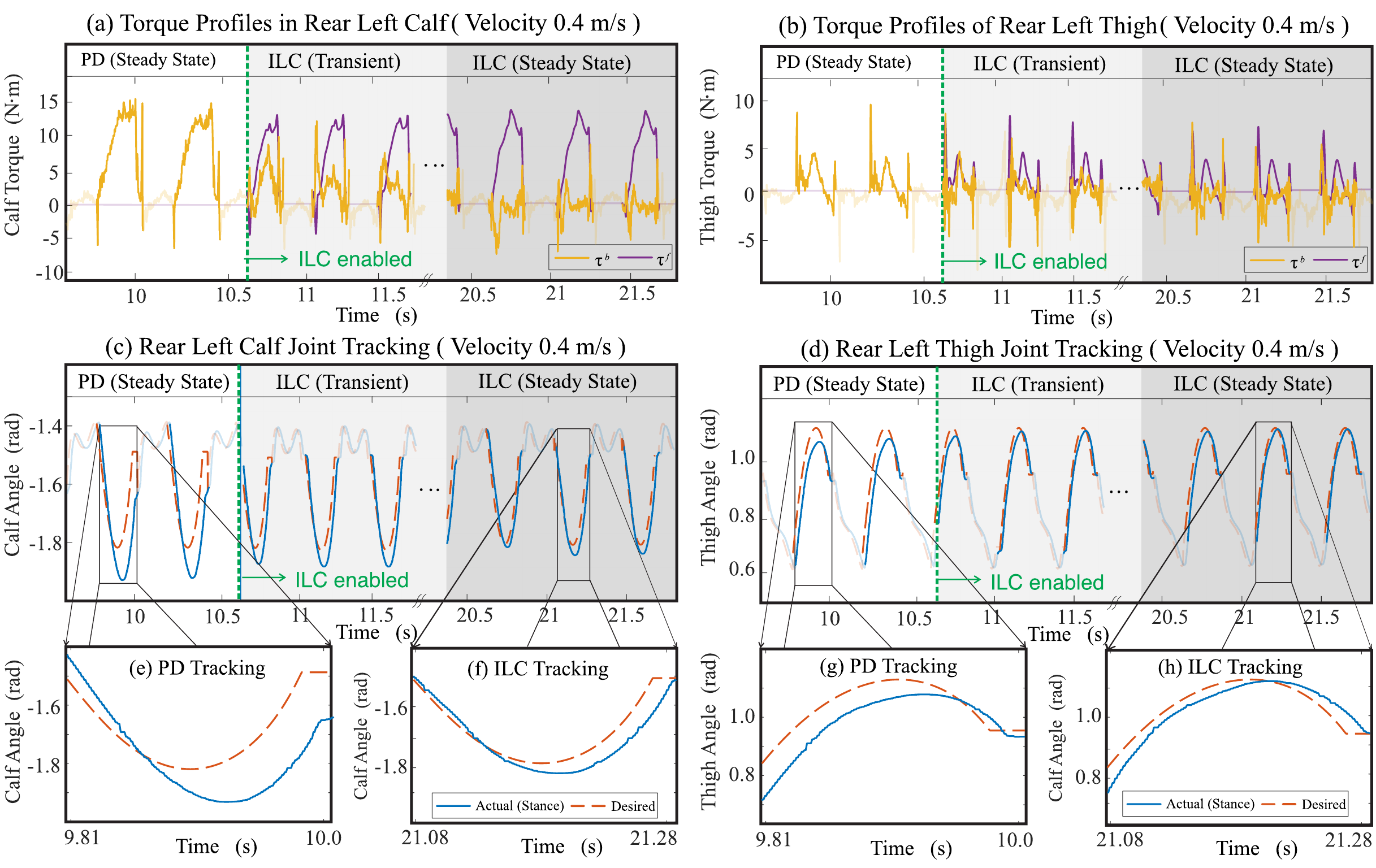}
\caption[ILC process]{Tracking performance of A1 during pronking at \SI[per-mode=symbol]{0.4}{\meter\per\second}. Subfigures (a)–(b) show control torque evolution before and after ILC activation at \( t = \SI{10.6}{\second} \). Subfigures (c)–(d) compare tracking errors under PD-only and ILC-based control, with calf and thigh RMSE reduced by 58.3\% and 25.0\%, respectively.}
\label{fig:calfandthigh}
\vspace{-2mm}
\end{figure*}

At a speed of \SI[per-mode=symbol]{0.4}{\meter\per\second}, Figure~\ref{fig:calfandthigh}(c) shows the tracking performance of the rear calf joint. Initially, under PD control, the joint exhibits noticeable errors, particularly during the mid-stance phase where gravitational demands are highest. The maximum tracking error reaches approximately \SI{0.26}{\radian}, indicating the PD controller's limited ability to compensate for unmodeled dynamics. After ILC is activated at \( t = \SI{10.6}{\second} \), the calf joint error reduces significantly. Within 12 strides (about \SI{5}{\second}), the error drops to \SI{0.14}{\radian}, and the RMSE improves from \SI{0.12}{\radian} to \SI{0.05}{\radian}, reflecting a 58.3\% reduction.
Figure~\ref{fig:calfandthigh}(d) illustrates the tracking performance of the rear thigh joint. The initial PD-controlled trajectory results in a peak error of \SI{0.15}{\radian}, especially prominent at the start of the stance phase due to fast loading. Once ILC is engaged, the error steadily decreases, and the RMSE improves from \SI{0.08}{\radian} to \SI{0.06}{\radian}, achieving a 25.0\% reduction. These results demonstrate that ILC effectively reduces tracking errors for both joints, with particularly strong improvements in joints subject to larger dynamic loads.

\begin{figure}[tbp]
\centering
\includegraphics[width=0.5\columnwidth]{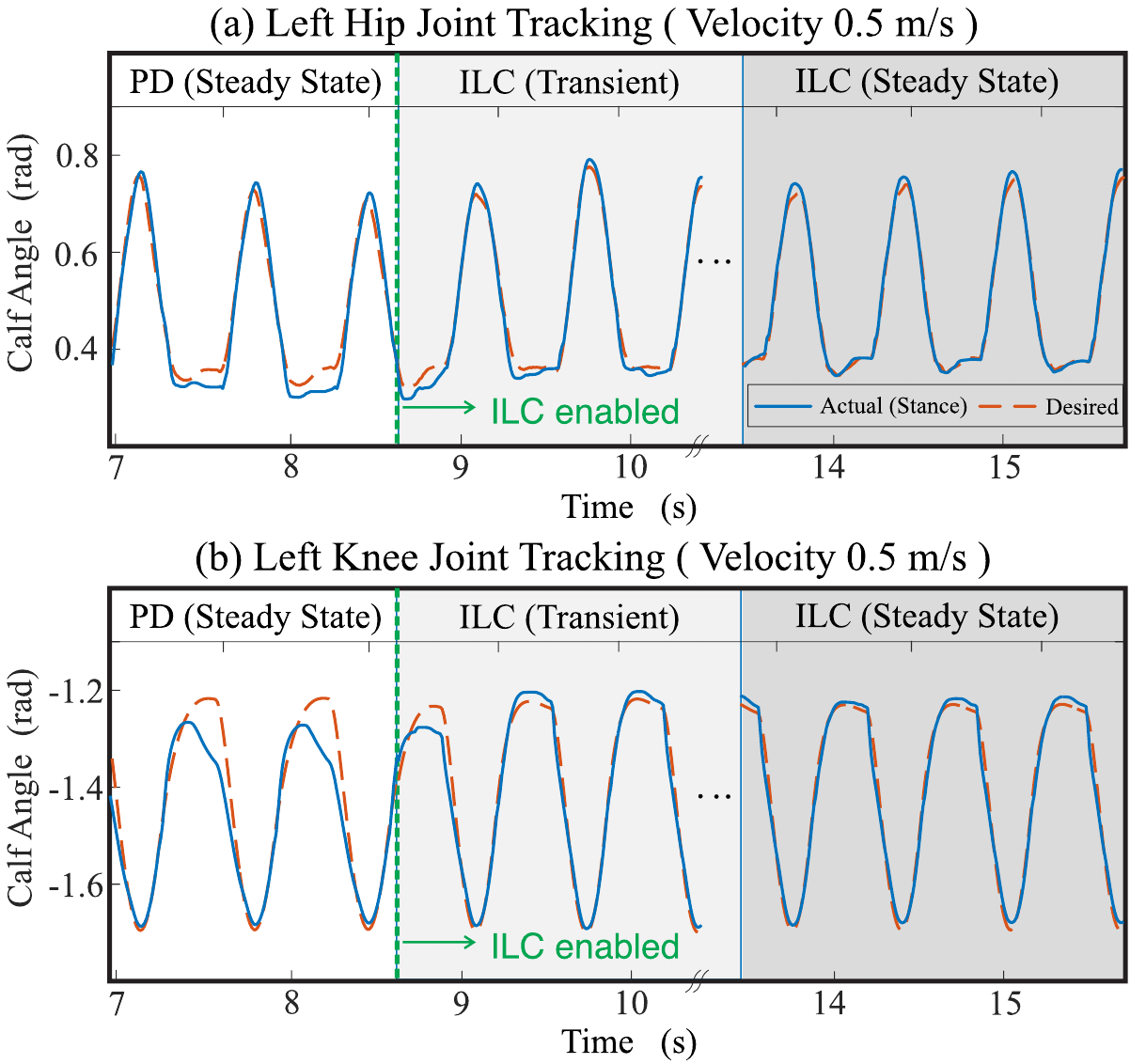}
\caption[Cassie tracking]{Hardware experiment on the Cassie robot showing tracking improvements for the left hip joint \( q_3 \) and knee joint \( q_4 \). The activation of ILC at \( t = \SI{8.6}{\second} \) resulted in up to 80\% reduction in tracking error within 3–5 strides.}
\label{fig:cassie_tracking}
\vspace{-2mm}
\end{figure}

To further validate the versatility of the proposed ILC framework, additional experiments were conducted on the Cassie bipedal robot. These experiments revealed even greater improvements in tracking performance compared to the A1 robot. The RMSE reductions for Cassie were consistently higher, with improvements exceeding 80\% for most joints. The results indicate that ILC can effectively adapt to different robotic platforms, further establishing its adaptability and scalability for advanced locomotion tasks.  Figure~\ref{fig:cassie_tracking} illustrates these results for the hip pitch and knee joints. Following ILC activation at \( t = \SI{8.6}{\second} \), tracking errors decreased significantly within just 3–5 strides, reinforcing the adaptability and scalability of the proposed learning framework across distinct robotic morphologies.

\subsubsection{ILC for Slope Ascension and Descension}

To assess the adaptability of the proposed controller under gravitational variations, hardware experiments were conducted on sloped surfaces ranging from \( -10^\circ \) (descending) to \( 10^\circ \) (ascending). These conditions introduce additional gravitational components along the direction of motion, increasing the challenge of maintaining precise trajectory tracking.
As shown in Table~\ref{table:2}, the ILC framework consistently improved tracking accuracy for both the calf and thigh joints across all slope angles. Specifically, RMSE reductions ranged from 50\% to 58\% for the calf joint and 20\% to 33\% for the thigh joint, when compared to baseline PD control. These results demonstrate the adaptability and generalizability of ILC in accommodating terrain-induced disturbances without requiring manual retuning of controller parameters.

\begin{table}[tbp]
\caption{Tracking performance improvements for the A1 robot on inclined and declined slopes. The RMSE reduction quantifies the improvement achieved by ILC relative to PD control.}
\label{table:2}
\setlength{\tabcolsep}{0.35em} 
\renewcommand{\arraystretch}{1.1}
\begin{center}
\vspace{-2mm}
\begin{tabular}{|c||c|c|r||c|c|r|}
\hline 
\multirow{2}{1cm}{\centering{\textbf{Slope (°)}}} & \multicolumn{3}{c||}{\textbf{Calf Error (rad)}} & \multicolumn{3}{c|}{\textbf{Thigh Error (rad)}} \\ \cline{2-7}
& \it PD & \it ILC & \it Reduction & \it PD & \it ILC & \it Reduction \\
\hline
$-10$ & 0.11 & 0.05 & 54.5\% & 0.05 & 0.04 & 20.0\% \\
$-7$  & 0.10 & 0.05 & 50.0\% & 0.06 & 0.04 & 33.3\% \\
$-5$  & 0.09 & 0.04 & 55.6\% & 0.06 & 0.04 & 33.3\% \\
$5$   & 0.12 & 0.06 & 50.0\% & 0.05 & 0.04 & 20.0\% \\
$7$   & 0.13 & 0.06 & 53.8\% & 0.07 & 0.05 & 28.6\% \\
$10$  & 0.12 & 0.05 & 58.3\% & 0.07 & 0.05 & 28.6\% \\
\hline
\end{tabular}
\vspace{-6mm}
\end{center}
\end{table}

\subsubsection{Generating the Torque Library}

The TL is constructed by extracting feedforward torque profiles from converged iterations of the ILC process. As learning progresses, the feedforward torques \( \vec{\tau}^{f}_{k}(s) \) gradually adapt to capture unmodeled dynamics, gravitational effects, and impact-related disturbances. Once convergence is achieved, the resulting torque profiles serve as reusable solutions for future executions of the same task.

Figure~\ref{fig:calfandthigh}(a) and (b) illustrate this process. Initially, the control input \( \vec{\tau}_{k}(s) \) relies heavily on feedback torques \( \vec{\tau}^{b}_{k}(s) \), which compensate for large tracking errors. As ILC iterates, the feedforward component \( \vec{\tau}^{f}_{k}(s) \) becomes increasingly dominant, leading to improved prediction and reduced reliance on feedback. At convergence, the change between successive feedforward commands becomes negligible, i.e., \( \vec{\tau}^{f}_{k}(s) \approx \vec{\tau}^{f}_{k-1}(s) \), indicating that the system has reached a steady-state solution.
In the figure, the converged feedforward torques after zero-phase filtering are shown in purple, while the residual feedback corrections are shown in yellow. Their sum constitutes the total joint torque \( \vec{\tau}_{k}(s) \), as defined in \eqref{eq:desiredtorque}. Once the convergence criterion in \eqref{eq:stop_condition} is satisfied, the torques from the final \( N_p = 12\) iterations 
strides are averaged and stored as the learned torque profile for the associated motion primitive $\vec{B}$.
This procedure is applied across all entries in the stored motion primitives, ensuring that each task is paired with an optimized torque profile $\vec{T}_{\tau}$. The resulting TL allows for fast initialization and execution of known motions, bypassing the need for repeated learning and improving both computational efficiency and adaptability in future deployments.

\subsubsection{Analyzing the Torque Library}
%
\begin{figure}[tb]
\centering
\includegraphics[width=0.5\columnwidth]{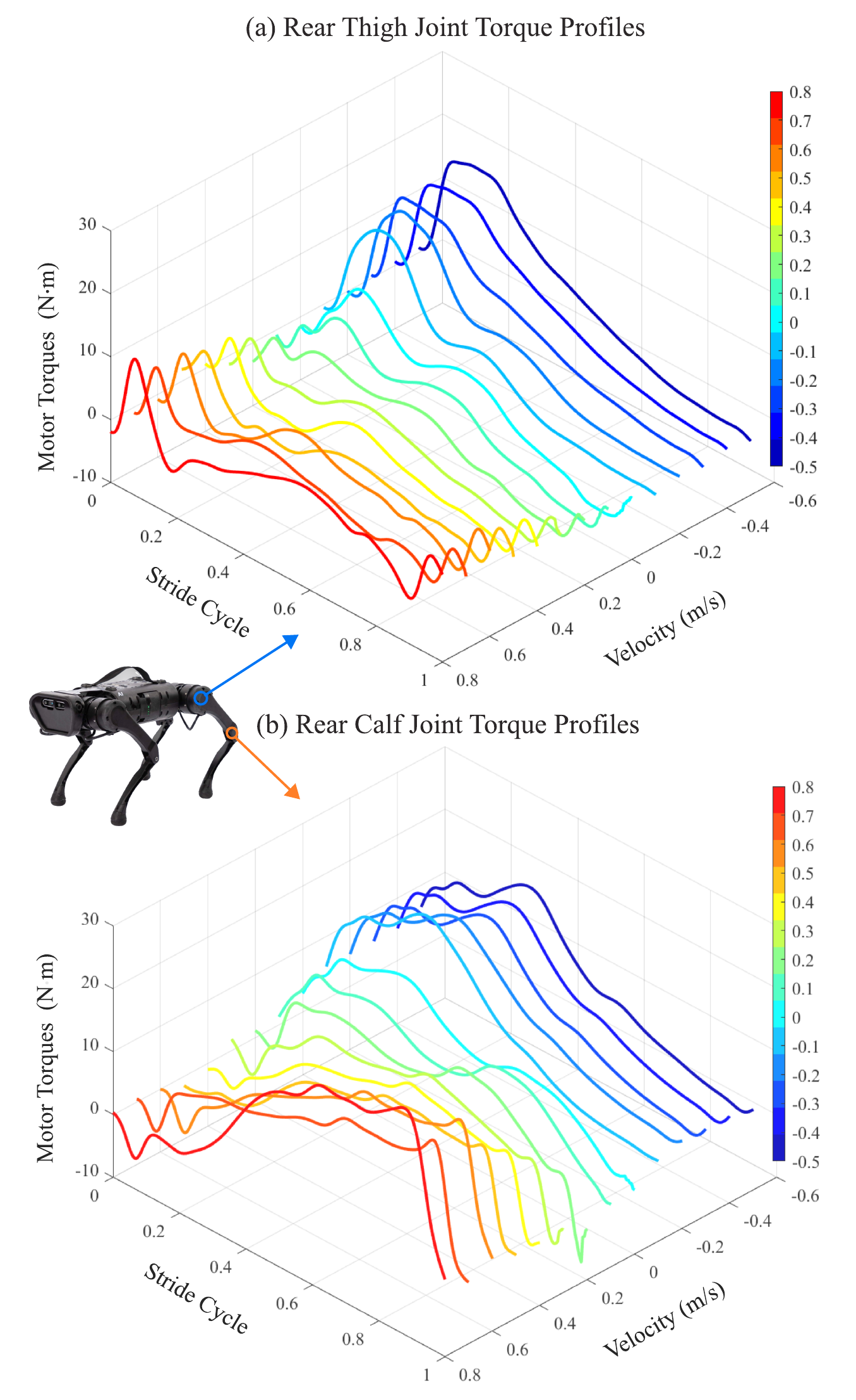}
\caption[Torque library]{Learned feedforward torque profiles for the A1 robot at various average speeds, organized within the TL. These profiles are directly used during online execution to provide predictive feedforward control without retraining. Panel (a) shows the rear thigh joint torque profiles, while panel (b) presents the rear calf joint. 
}
\label{fig:TL}
\end{figure} 

%
\begin{figure}[tbp]
\centering
\includegraphics[width=0.5\columnwidth]{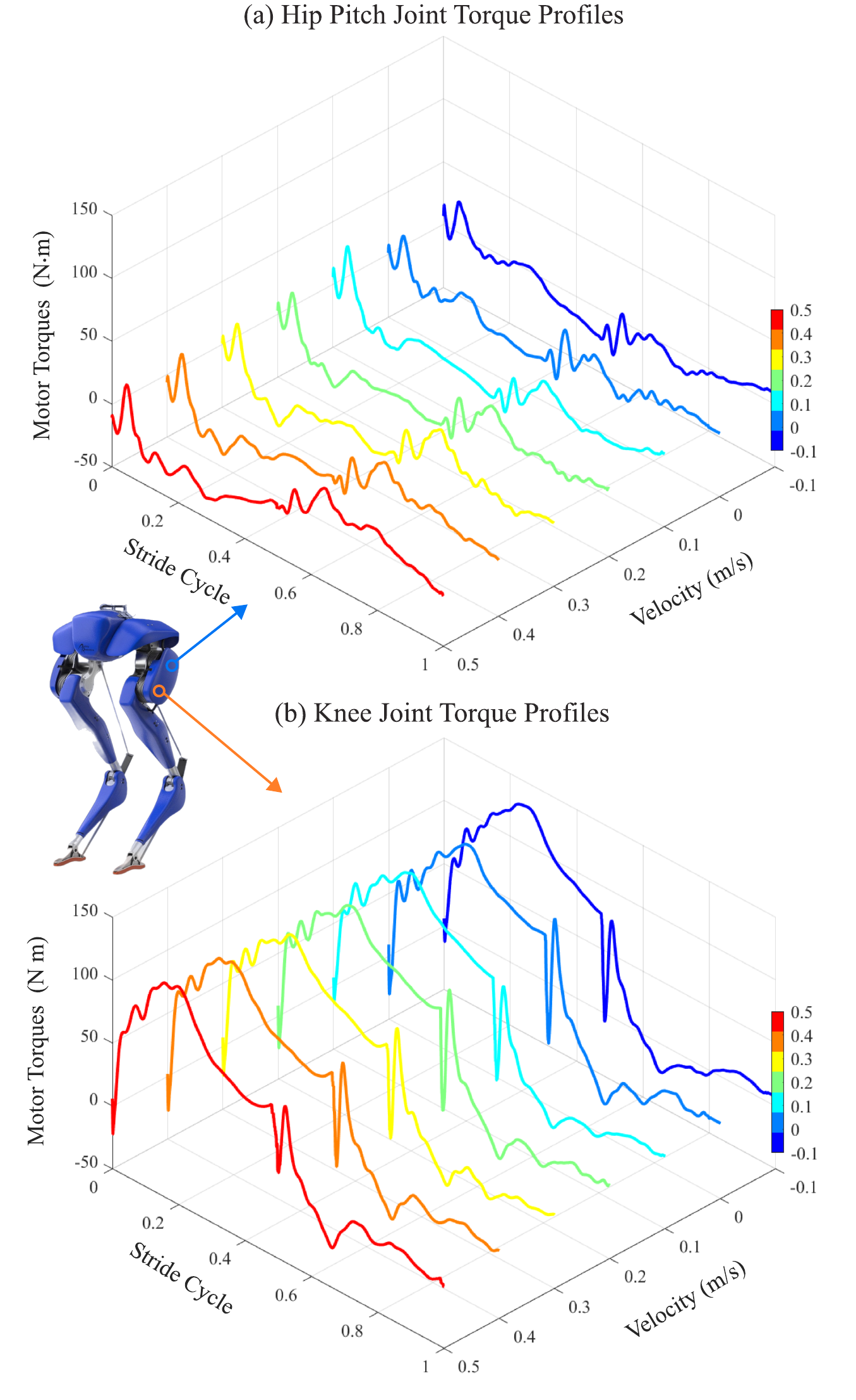}
\caption[Cassie torque library]{Construction of the TL for Cassie’s left leg. Each black line represents the average feedforward torque trajectory derived from 20 ILC-converged trials at a given speed. Panel (a) displays the hip pitch joint ($q_3$), and panel (b) shows the knee joint ($q_4$). 
}
\label{fig:cassie_torques}
\vspace{-2mm}
\end{figure}
%
\begin{figure}[tbp]
\centering
\includegraphics[width=0.5\columnwidth]{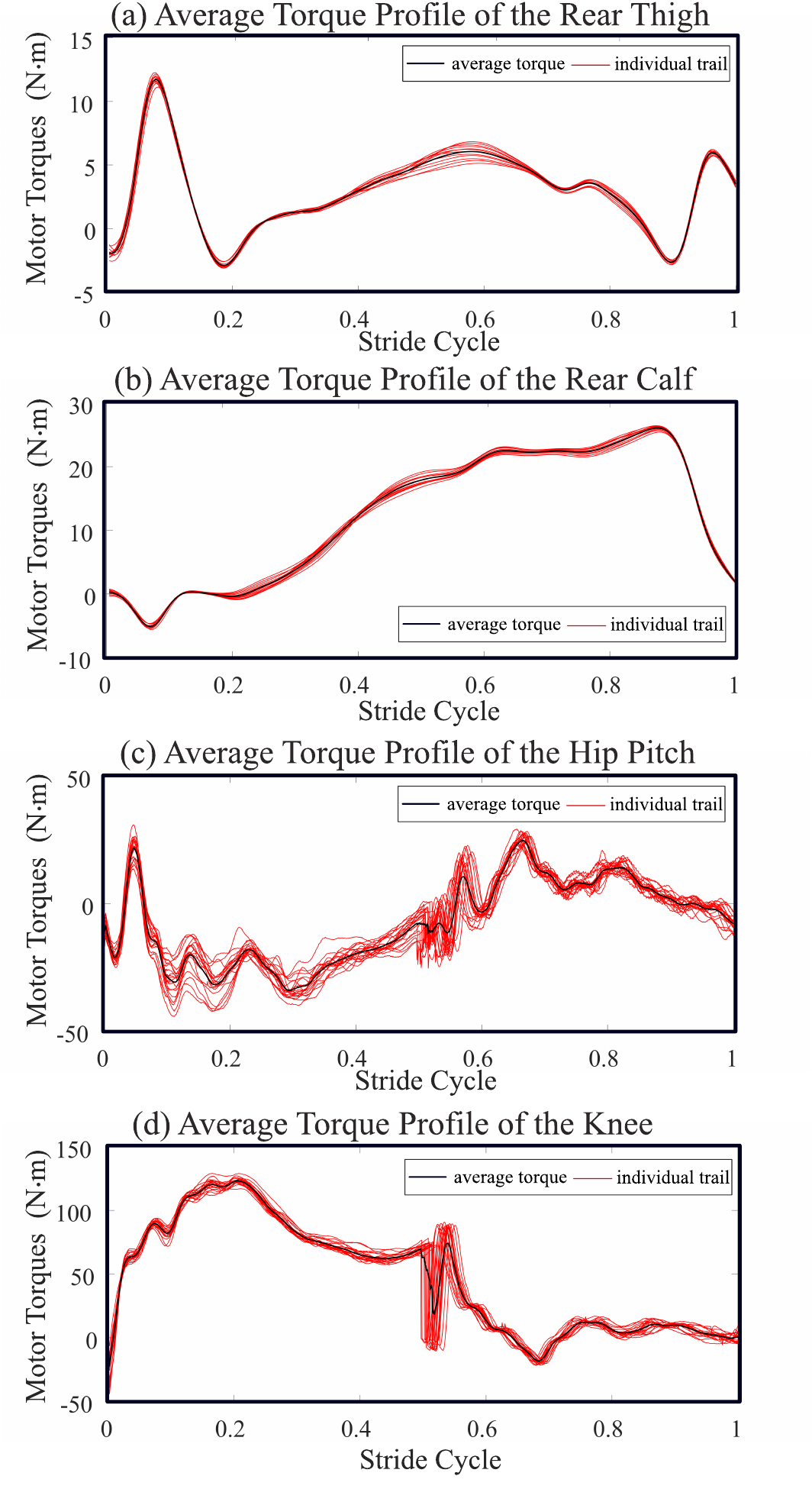}
\caption[The average torque profiles]{The average torque profiles at \SI[per-mode=symbol]{0.5}{\meter\per\second} for A1 model and \SI[per-mode=symbol]{0.8}{\meter\per\second} for Cassie model alongside multiple trial curves for: (a) rear thigh joint in A1, (b) rear calf joint in A1, (c) hip pitch joint in Cassie, and (d) knee joint in Cassie. (a) and (b) present the averaged torque profiles computed across 12 trials, while (c) and (d) display the averaged results from 20 trials. Individual trial curves are shown in the red to illustrate variability, with the mean curve highlighting in black.}
\label{fig:average}
\vspace{-2mm}
\end{figure}
The TL organizes the converged feedforward torque profiles into a structured lookup table indexed by task parameters such as velocity. Figure~\ref{fig:TL} visualizes a subset of the TL constructed for the A1 robot, showing torque profiles learned for trajectories with average speeds ranging from \SI[per-mode=symbol]{-0.6}{\meter\per\second} to \SI[per-mode=symbol]{0.8}{\meter\per\second}, in increments of \SI[per-mode=symbol]{0.1}{\meter\per\second}, the stride cycle is defined as the time between two successive ground contact events. These profiles are directly deployed during online control, eliminating the need for real-time adaptation or re-learning.
Distinct trends emerge when comparing forward and backward locomotion. During forward motion, the rear calf joint torque increases sharply at the start of the stance phase, peaks near mid-stance, and then declines. In contrast, backward motion induces an early peak due to elevated loading, followed by a second rise near mid-stance. The rear thigh joint torque shows a gradual increase that peaks just before liftoff, reflecting the transition into the flight phase. Notably, the rear thigh exhibits similar qualitative trends in both directions, highlighting its central role in velocity regulation. These variations align with mechanical asymmetries in the robot's front-rear leg design.
Despite the differences in magnitude and timing, the overall torque profile shapes remain consistent across joints, suggesting strong generalizability. The TL’s coverage of directional and velocity-dependent variations ensures adaptability to a wide range of locomotion scenarios. The hybrid controller, combined with progressive interpolation in the phase space, allows smooth application of these profiles even when intermediate velocities are encountered.

Figure~\ref{fig:cassie_torques} shows the corresponding TL construction for the Cassie robot's left knee joint. Here, the TL is generated for average speeds between \SI[per-mode=symbol]{-0.1}{\meter\per\second} and \SI[per-mode=symbol]{0.5}{\meter\per\second}. At each speed, 20 converged torque profiles are collected, filtered, and averaged to form a representative entry in the library. 
Figure~\ref{fig:average} illustrates the averaging process across 12 trials for the A1 robot and 20 trials for the Cassie robot. The black curves in Figure~\ref{fig:average}(a) and (b) represent the average torque profiles of the thigh and calf joints of the A1 robot, corresponding to A1 pronking at \SI[per-mode=symbol]{0.8}{\meter\per\second}, as shown in Figure~\ref{fig:TL}. Similarly, the black curves in Figure~\ref{fig:average}(c) and (d) show the average torque profiles of the hip pitch and knee joints of the Cassie robot, corresponding to Cassie walking at \SI[per-mode=symbol]{0.5}{\meter\per\second}, as depicted in Figure~\ref{fig:cassie_torques}.

\begin{figure}[tbp]
\centering
\includegraphics[width=0.5\columnwidth]{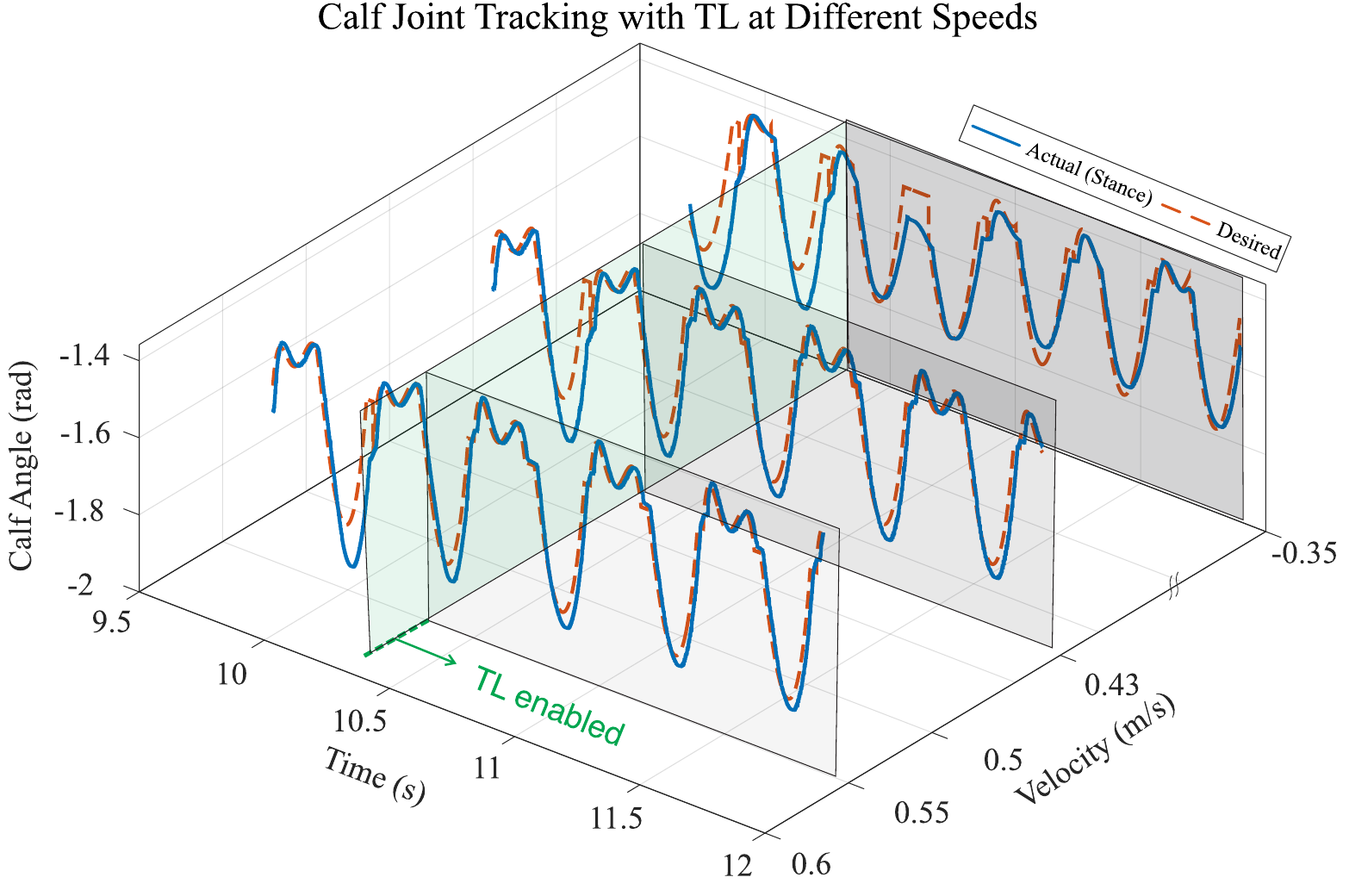}
\caption[Simulation results]{Tracking performance of the A1 robot at interpolated speeds (\SI[per-mode=symbol]{-0.35}{\meter\per\second}, \SI[per-mode=symbol]{0.43}{\meter\per\second}, and \SI[per-mode=symbol]{0.55}{\meter\per\second}) using the TL. The interpolated feedforward torques significantly reduced convergence time, with the robot achieving steady-state tracking within just two strides. 
}
\label{fig:withTL}
\vspace{-2mm}
\end{figure}

The TL enables the robot to perform locomotion at arbitrary speeds by linearly interpolating between stored torque profiles using \eqref{eq:TL_interp}. This interpolation ensures smooth transitions in feedforward compensation as the torso velocity changes, maintaining consistent tracking performance without requiring additional learning iterations.
In practical evaluations, the robot achieved rapid convergence to target speeds within approximately \SI{1.5}{\second}, a substantial improvement over the initial ILC learning duration of about \SI{5}{\second} (12 strides), as shown in Figure~\ref{fig:calfandthigh}. Figure~\ref{fig:withTL} illustrates the tracking performance at interpolated speeds of \SI[per-mode=symbol]{-0.35}{\meter\per\second}, \SI[per-mode=symbol]{0.43}{\meter\per\second}, and \SI[per-mode=symbol]{0.55}{\meter\per\second}. In all cases, stable tracking was achieved within just two strides, demonstrating the TL’s effectiveness in enabling fast adaptation and enhancing overall control efficiency.

\subsubsection{Comparison with Whole-Body Control}

\begin{figure}[tbp]
\centering
\includegraphics[width=0.5\columnwidth]{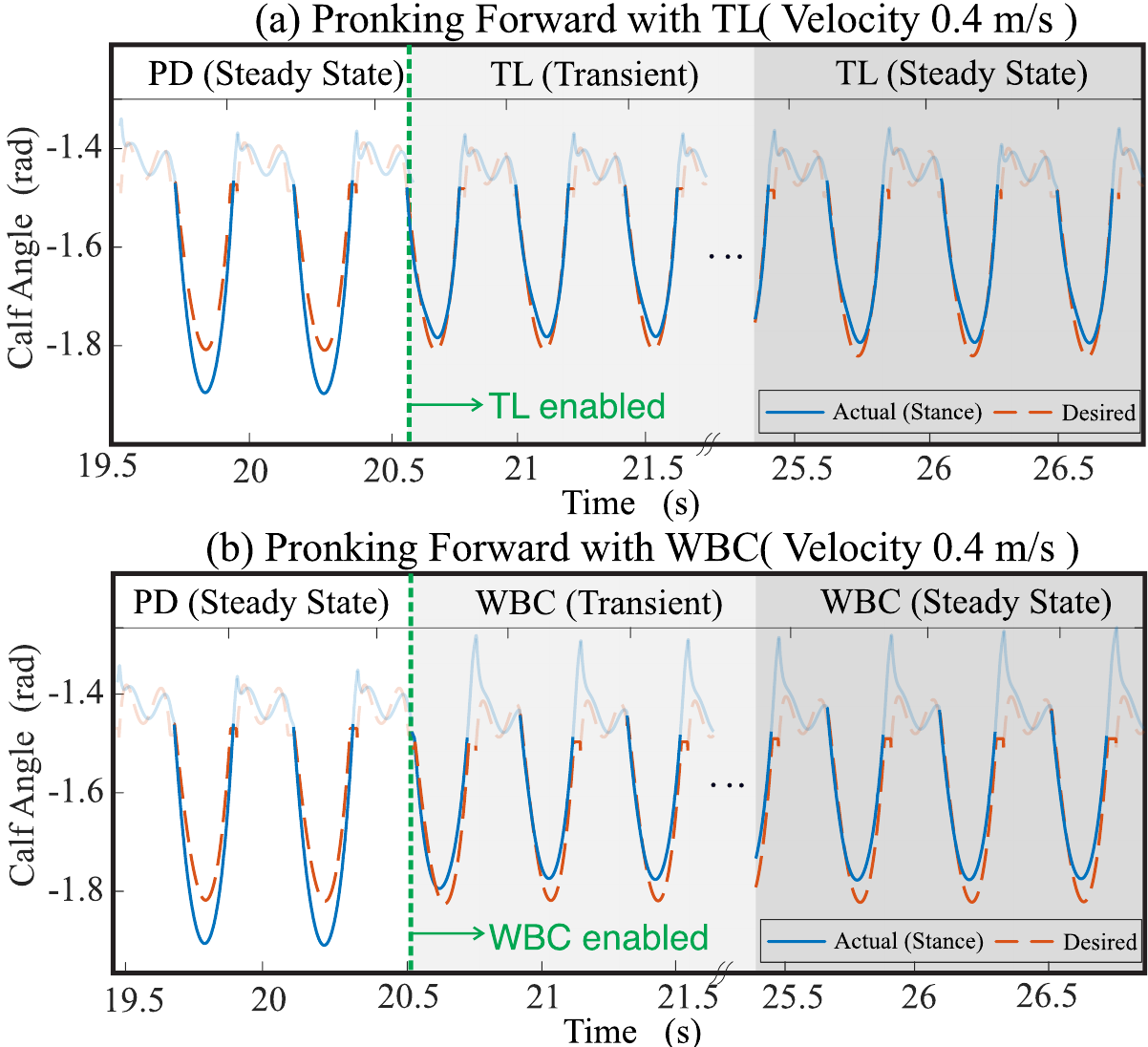}
\caption[Simulation results]{\textbf{Gazebo simulation comparison} of calf joint tracking performance: (a) TL-based control; (b) WBC. The TL exhibits improved accuracy and reduced overshoot compared to WBC. The activation of TL at \( t = \SI{20.57}{\second} \) resulted in 71.7\% RMSE reduction in calf. The activation of WBC at \( t = \SI{20.51}{\second} \) resulted in 48.3\% RMSE reduction in calf.
}
\label{fig:TLvsWBCsim}
\vspace{-2mm}
\end{figure}

\begin{figure}[tbp]
\centering
\includegraphics[width=0.5\columnwidth]{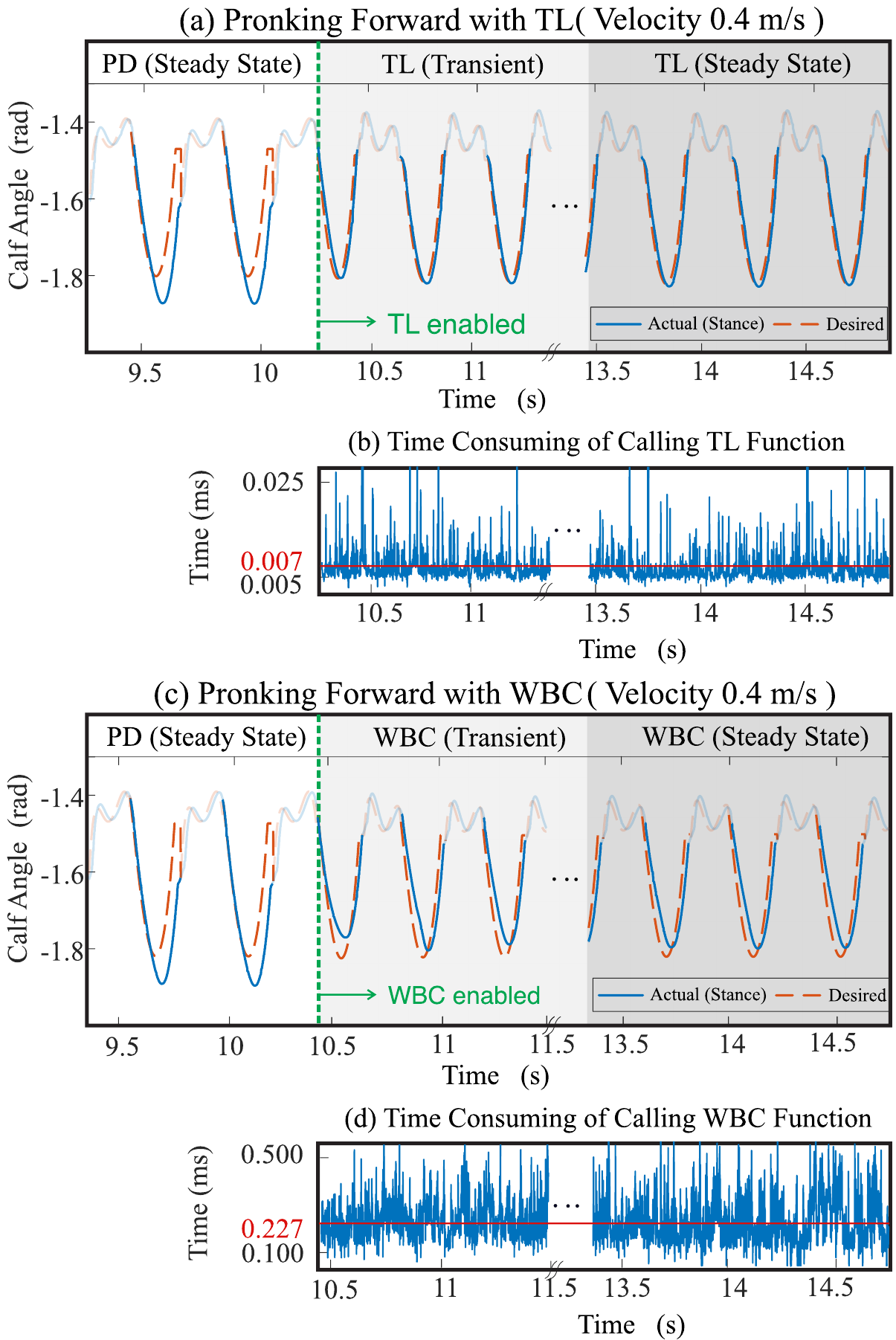}
\caption[Hardware test results]{A1 hardware comparison of tracking performance and computation time: (a) Tracking using the TL-based controller with 57.7\% RMSE reduction; (c) Tracking using WBC with 50.4\% RMSE reduction. (b) TL function call time: \( 0.0065 \, \si{ms} \); (d) WBC function call time: \( 0.2274 \, \si{ms} \). The TL-based controller not only delivers \textbf{over 35 times faster} computation but also achieves \textbf{better trajectory tracking} than WBC, making it highly suitable for real-time, high-performance locomotion. 
}
\label{fig:TLvsWBC}
\vspace{-2mm}
\end{figure}

This section compares the TL-based control strategy with the conventional WBC approach in both simulation and hardware experiments. Figure~\ref{fig:TLvsWBCsim} presents Gazebo simulation results for the calf joint, while Figure~\ref{fig:TLvsWBC} shows corresponding tracking and computation performance on the A1 robot.

While both methods achieve basic trajectory tracking, WBC exhibits overcompensation and increased steady-state error due to its reliance on accurate dynamic modeling. This dependency makes WBC computationally demanding and less robust in real-world conditions, where modeling inaccuracies and external disturbances are common. As shown in Figure~\ref{fig:TLvsWBC}(d), WBC requires approximately \SI{0.2274}{\milli\second} per control iteration, which poses a challenge for maintaining high-frequency updates in dynamic environments.

In contrast, the TL-based approach benefits from the data-driven refinement enabled by ILC. Once learned, the feedforward torque profiles stored in the TL capture the dominant task dynamics, including unmodeled effects such as friction, joint backlash, and terrain variability. This eliminates the need for repeated inverse dynamics computations, significantly reducing runtime. As shown in Figure~\ref{fig:TLvsWBC}(b), the TL-based controller achieves an average execution time of only \SI{0.0065}{\milli\second}, enabling real-time deployment with low latency. 

These comparisons demonstrate that \textbf{the TL-based control strategy not only achieves superior tracking accuracy but also improves computational efficiency by more than 35 times, making it highly suitable for real-time execution in fast, agile, and adaptive locomotion tasks.} The learned torque trajectories stored in the TL encode task-specific actuation strategies, effectively serving as reusable ``motor memories'' for the robot. \textbf{Once embedded in the TL, these profiles eliminate the need for repeated learning or real-time inverse dynamics computation, enabling rapid task execution with high precision and minimal computational overhead.} 
Moreover, the stored profiles generalize effectively across diverse locomotion scenarios, accommodating variations in speed, terrain, gait type, and robot morphology without requiring manual retuning. By decoupling feedforward generation from online optimization, the proposed approach achieves adaptive, data-efficient control while offering a scalable framework applicable to both quadrupedal and bipedal robots.

\subsection{Nonperiodic Motion with Learned Torque}

The preceding simulation and experimental results confirm that the proposed ILC-based control method is particularly effective for (quasi-)periodic tasks by leveraging the temporal continuity inherent in repetitive motions. In such scenarios, the robot rapidly refines its control strategy and establishes task-specific torque profiles within a few iterations. 
However, for nonperiodic or single-shot motions such as rapid maneuvers including jumps or flips repetition opportunities are limited. These motions do not occur cyclically, and the same control instruction may not be executed again until a similar task is encountered. To address this, the framework includes an online data logging mechanism that captures control signals and tracking errors in real time. This allows the feedforward torque commands learned during rare or nonrepeating events to be stored and reused when the same motion is attempted again. 
While learning in these cases is temporally sparse and discrete, the framework’s rapid convergence properties allow effective “motor memories” to form from limited data. This capability makes the approach well suited for a broad range of agile, high-impact motions without requiring extensive retraining or large datasets. Moreover, the proposed approach enables the robot to interactively practice and master individual skills through repeated trials. Once a specific motion is successfully learned, the resulting torque profile can be reused to warm-start the training of more challenging tasks that build upon the same dynamics. This hierarchical skill composition accelerates learning and facilitates progressive generalization to increasingly complex behaviors.

%
\begin{figure}[tbp]
\centering
\includegraphics[width=0.5\columnwidth]{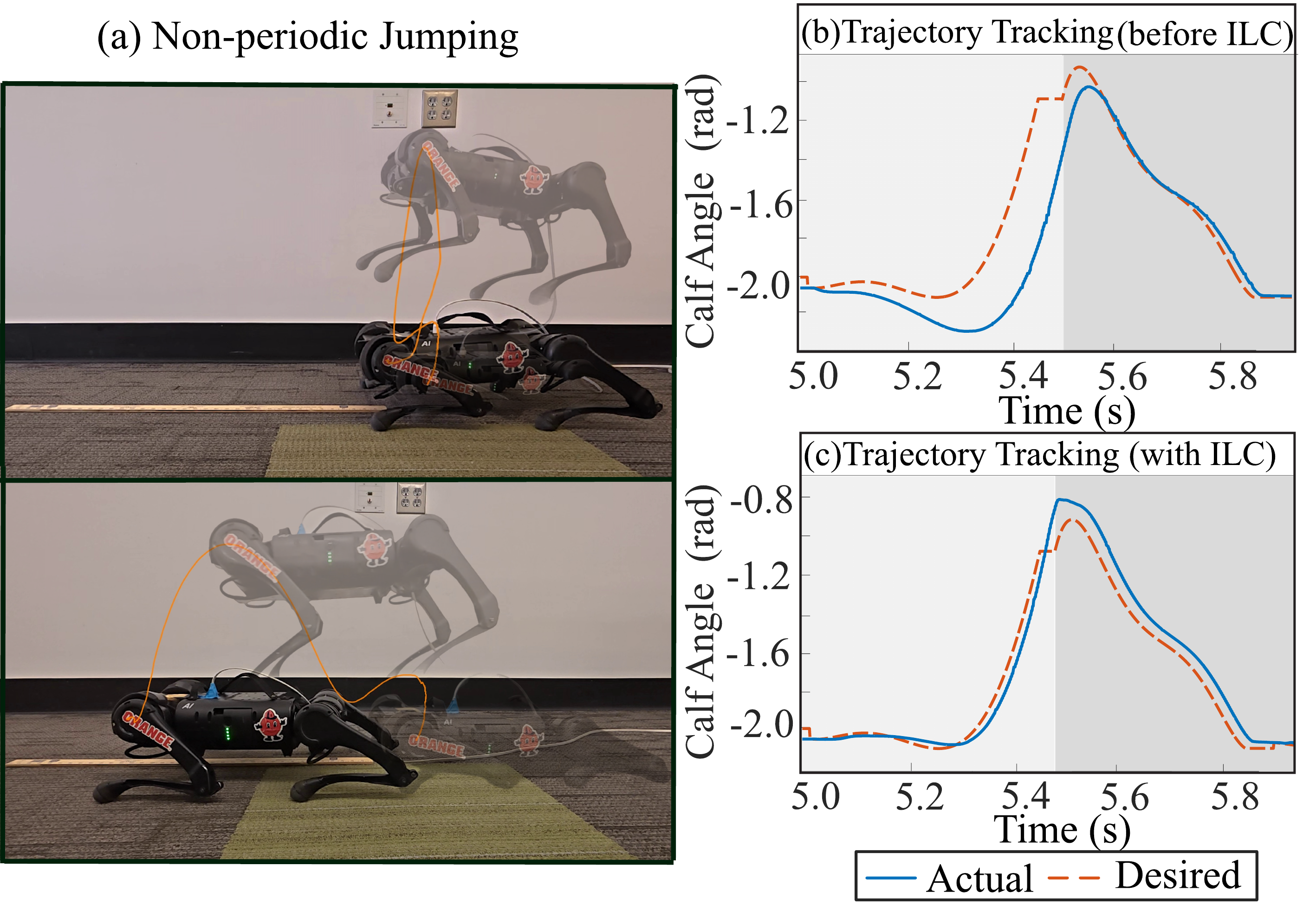}
\caption[Jumping]{Improvement in jumping performance with learned torque. (a) Jumping distance before learning is approximately \SI{0.09}{\meter}; (b) calf joint tracking prior to ILC; (c) improved tracking after ILC results in a final jump distance of \SI{0.39}{\meter}.
}
\label{fig:Jumping}
\vspace{-2mm}
\end{figure}
%

To demonstrate this capability, we evaluate a discrete jumping task. In this experiment, the robot is commanded to achieve a horizontal jump of \SI{0.35}{\meter}. As shown in Figure~\ref{fig:Jumping}(a), the jump distance before learning is only \SI{0.09}{\meter}, verified through visual analysis of recorded video footage. 
Figure~\ref{fig:Jumping}(b) presents the corresponding calf joint tracking performance. The light-shaded region denotes the stance phase following the preparatory posture, while the dark-shaded region spans the flight phase and subsequent touchdown.
After applying the ILC process over three iterations, the controller significantly improves tracking performance, as illustrated in Figure~\ref{fig:Jumping}(c). The refined feedforward torque profiles enable accurate compensation for unmodeled dynamics, allowing the robot to achieve the target jump distance of \SI{0.39}{\meter}. The actual and desired motions of the hip joint are shown in the blue line and the orange dashed line, respectively, in Figure~\ref{fig:Jumping}(a).
Moreover, the jump distance can be modulated by adjusting the torque profile at the thigh joint, analogous to the foot placement strategy used for speed control in Raibert’s work \citep[Chapter~3]{raibert1986legged}. These findings demonstrate that even single-execution motions can leverage the learned torque strategy, contributing to the construction of a TL that enhances landing precision and consistency across diverse environmental conditions.

\section{Discussion and Limitation}
\label{sec:discussion}

The proposed hybrid control framework combines ILC with a TL to deliver a scalable, efficient, and adaptive locomotion solution for legged robots. Compared to model-based methods such as WBC, which depend on accurate dynamics and incur high online computation, our approach refines feedforward torques directly from execution and reuses them without runtime optimization. This results in control update rates over 30$\times$ faster than conventional controllers, enabling fast, reliable operation in real-world settings.
The framework offers several key advantages:  
\textbf{1) Rapid learning from hardware interaction:} Leveraging the repetitive nature of locomotion, the robot can refine its control strategy in just a few iterations and achieve up to 85\% error reduction within seconds.  
\textbf{2) Generalization via torque memory:} Learned torque profiles are stored in the TL and interpolated to adapt to variations in speed, terrain, and gravity, allowing the robot to perform new tasks without retraining.  
\textbf{3) Hierarchical skill acquisition:} Once a simple task is learned, its torque profile can warm-start more complex motions, enabling efficient curriculum-style learning and progressive skill composition.
The experiments on both the quadrupedal A1 robot and the bipedal Cassie robot highlight the framework’s scalability and versatility. While prior work has often focused on specific platforms or gaits \cite{bledt2018cheetah,mordatch2010robust}, our results demonstrate consistent performance across different morphologies, speeds, and terrains. The ability to reach steady-state tracking within a few strides and maintain low RMSE across diverse tasks underscores the adaptability of the proposed approach. Furthermore, the TL-based controller consistently outperforms WBC in steady-state error reduction, illustrating the limitations of model-based strategies in the presence of unmodeled dynamics.

Despite these advantages, several limitations remain. In its current implementation, the learning performance is sensitive to the design of the zero-phase filters described in Section~\ref{sec:filter}, which affect convergence speed and stability. However, this limitation can potentially be addressed by incorporating predictive filtering or trajectory forecasting strategies inspired by MPC, as explored in prior work \cite{nguyen2024mastering}. Additionally, the use of linear interpolation within the TL may restrict generalization accuracy in highly nonlinear or high-dimensional task spaces. Extending the TL with richer interpolation schemes or task-conditioned embeddings could improve adaptability. Finally, although the method has been validated across multiple robots and environments, further work is needed to generalize it to hybrid behaviors such as manipulation or whole-body dynamic tasks.
Nevertheless, the proposed method offers a data-efficient, computation-light, and generalizable approach to legged locomotion. By unifying iterative learning with reusable motor memories, it provides a practical and robust alternative to conventional control architectures for adaptive, high-performance mobility in real-world conditions.

\section{Conclusion}
\label{sec:conclusion}
To achieve reliable, high-performance locomotion in unstructured environments, legged robots must be capable of rapidly adapting to changing dynamics without relying on precise models or extensive computation. In this paper, we presented a hybrid control framework that integrates ILC with a biologically inspired TL to improve motion accuracy, adaptability, and efficiency in both periodic and nonperiodic tasks. 
We demonstrated that the ILC-based controller can refine feedforward torque profiles directly from execution and store these profiles in a TL, enabling fast reuse across tasks and eliminating the need for online optimization. Extensive simulations and hardware experiments were conducted on both the A1 quadrupedal robot and the Cassie biped, showing consistent performance across diverse terrains, speeds, and gravitational conditions. In these experiments, the controller achieved up to 85\% reduction in joint tracking error within a few strides. Furthermore, we showed that the TL enables smooth generalization via interpolation and supports hierarchical skill acquisition by allowing learned motions to warm-start more complex behaviors.

Future work will focus on generalizing the proposed method to a broader class of locomotion and interaction tasks, including contact-rich manipulation, whole-body multi-contact behaviors, and variable morphology systems. To support this, we plan to explore ILC formulations in the frequency domain, which may improve convergence and robustness in more complex or non-repetitive motion patterns. We also intend to investigate predictive adaptation strategies and the use of task-conditioned torque embeddings, enabling more expressive and flexible representation within the TL. Finally, we will examine how the resolution and structure of the TL affect generalization in high-dimensional task spaces and validate the framework across new robotic morphologies and task domains.



\begin{con}
\ctitle{Acknowledgements}
The authors would like to thank Dr. Jessy W. Grizzle and Dr. Ram Vasudevan from the University of Michigan for providing access to the Cassie Blue and Cassie Maize robots, as well as the associated experimental facilities. Their generous support and resources were instrumental in enabling the successful execution of the experiments presented in this work.

\ctitle{Author Contributions}
Jing Cheng conceptualized the study, with guidance from Zhenyu Gan, and defined the research objectives. Jing Cheng conducted the simulations, developed the A1 robot experimental protocol, and wrote the initial draft of the manuscript. Zhenyu Gan contributed to the literature review, data analysis, and Cassie robot experiments, and provided oversight and guidance for research planning and execution. Yasser G. Alqaham contributed to the development of the gait library and assisted with the experiments. All authors discussed the results, reviewed the manuscript, and approved the final version.

\ctitle{Financial Support}
This work was supported in part by the National Science Foundation under Award Nos.~2525502 and 2315695.

\ctitle{Conflicts of Interest}
 The authors declare no conflicts of interest exist.

\ctitle{Ethical Approval}
Not applicable.

\end{con}

\vspace{5mm} 


\begin{thebibliography}{99}

\bibitem{unitree2020a1}
{Unitree Robotics} 2020.
\newblock {Unitree A1 Quadruped Robot}.
\newblock \url{https://www.unitree.com/a1/}.
\newblock Accessed: 2025-03-29.

\bibitem{agilityrobotics_cassie}
{Agility Robotics} 2025.
\newblock {Cassie Bipedal Robot}.
\newblock \url{https://agilityrobotics.com/}.
\newblock Accessed: 2025-03-30.

\bibitem{kim2019highly}
D. Kim, J.~D. Carlo, B. Katz, G. Bledt, and S. Kim 2019.
\newblock Highly dynamic quadruped locomotion via whole-body impulse control
  and model predictive control.
\newblock arXiv:1909.06586.

\bibitem{Katz_2019}
B. Katz, J.~D. Carlo, and S. Kim.
\newblock Mini Cheetah: A platform for pushing the limits of dynamic quadruped
  control.
\newblock In {\em 2019 International Conference on Robotics and Automation
  (ICRA)} 2019, pp. 6295--6301.

\bibitem{kajita2001simple}
S. Kajita, F. Kanehiro, K. Kaneko, K. Yokoi, and H. Hirukawa.
\newblock The {3D} linear inverted pendulum mode: A simple modeling for a
  biped walking pattern generation.
\newblock In {\em Proceedings of the IEEE/RSJ International Conference on
  Intelligent Robots and Systems (IROS)} 2001, pp. 239--246. IEEE.

\bibitem{wieber2006trajectory}
P.-B. Wiebar.
\newblock Trajectory free linear model predictive control for stable walking
  in the presence of strong perturbations.
\newblock In {\em 2006 6th IEEE-RAS International Conference on Humanoid
  Robots} 2006, pp. 137--142.

\bibitem{neunert2018whole}
M. Neunert, F. Farshidian, M. Wermelinger, A. St{\"a}uble, and J. Buchli.
\newblock Whole-body nonlinear model predictive control through contacts for
  quadrupeds.
\newblock {\em IEEE Robotics and Automation Letters}, {\bf 3}(3), 1458--1465
  (2018).


\bibitem{kang2022nonlinearMPC}
D. Kang, F.~D. Vincenti, and S. Coros 2022.
\newblock Nonlinear model predictive control for quadrupedal locomotion using
  second-order sensitivity analysis.
\newblock arXiv:2207.10465.

 
\bibitem{hwangbo2020learning}
J. Lee, J. Hwangbo, L. Wellhausen, V. Koltun, and M. Hutter.
\newblock Learning quadrupedal locomotion over challenging terrain.
\newblock {\em Science Robotics}, {\bf 5}(47), eabc5986 (2020).
 
\bibitem{siekmann2021blind}
J. Siekmann, K. Green, J. Warila, A. Fern, and J. Hurst 2021.
\newblock Blind bipedal stair traversal via sim-to-real reinforcement learning.
\newblock arXiv:2105.08328.

 
\bibitem{gu2017deep}
S. Gu, E. Holly, T. Lillicrap, and S. Levine.
\newblock Deep reinforcement learning for robotic manipulation with
  asynchronous off-policy updates.
\newblock In {\em 2017 IEEE International Conference on Robotics and Automation
  (ICRA)} 2017, pp. 3389--3396.
 
\bibitem{haarnoja2018soft}
T. Haarnoja, A. Zhou, P. Abbeel, and S. Levine.
\newblock Soft actor-critic: Off-policy maximum entropy deep reinforcement
  learning with a stochastic actor.
\newblock In {\em Proceedings of the 35th International Conference on Machine
  Learning (ICML)} 2018, volume~80, pp. 1861--1870. PMLR.
 
\bibitem{peng2018sim}
X. Peng, M. Andrychowicz, W. Zaremba, and P. Abbeel.
\newblock Sim-to-real transfer of robotic control with dynamics randomization.
\newblock In {\em 2018 IEEE International Conference on Robotics and Automation
  (ICRA)} 2018, pp. 3803--3810. IEEE.
 
\bibitem{zhao2020sim}
W. Zhao, J.~P. Queralta, and T. Westerlund.
\newblock Sim-to-real transfer in deep reinforcement learning for robotics:
  a survey.
\newblock In {\em 2020 IEEE Symposium Series on Computational Intelligence
  (SSCI)} 2020, pp. 737--744.


\bibitem{hansen2021}
N. Hansen, R. Jangir, Y. Sun, G. Aleny\`{a}, P. Abbeel, A.~A. Efros,
  L. Pinto, and X. Wang 2021.
\newblock Self-supervised policy adaptation during deployment.
\newblock arXiv:2007.04309.
 
\bibitem{yang2022RL}
T.-Y. Yang, T. Zhang, L. Luu, S. Ha, J. Tan, and W. Yu 2022.
\newblock Safe reinforcement learning for legged locomotion.
\newblock arXiv:2203.02638.
 
\bibitem{ahn2007iterative}
H.-S. Ahn, Y. Chen, and K.~L. Moore.
\newblock Iterative learning control: Brief survey and categorization.
\newblock {\em IEEE Transactions on Systems, Man, and Cybernetics, Part C
  (Applications and Reviews)}, {\bf 37}(6), 1099--1121 (2007).
 
\bibitem{bristow2006survey}
D.~A. Bristow, M. Tharayil, and A.~G. Alleyne.
\newblock A survey of iterative learning control.
\newblock {\em IEEE Control Systems Magazine}, {\bf 26}(3), 96--114 (2006).
 
\bibitem{cheng2023practice}
J. Cheng, Y.~G. Alqaham, A.~K. Sanyal, and Z. Gan.
\newblock Practice makes perfect: an iterative approach to achieve precise
  tracking for legged robots.
\newblock In {\em 2023 American Control Conference (ACC)} 2023, pp. 2165--2170.
 
\bibitem{di2018dynamic}
J. Di~Carlo, P.~M. Wensing, B. Katz, G. Bledt, and S. Kim.
\newblock Dynamic locomotion in the {MIT} Cheetah 3 through convex
  model-predictive control.
\newblock In {\em 2018 IEEE/RSJ International Conference on Intelligent Robots
  and Systems (IROS)} 2018, pp. 1--9.

\bibitem{9196562}
O. Melon, M. Geisert, D. Surovik, I. Havoutis, and M. Fallon.
\newblock Reliable trajectories for dynamic quadrupeds using analytical costs
  and learned initializations.
\newblock In {\em 2020 IEEE International Conference on Robotics and Automation
  (ICRA)} 2020, pp. 1410--1416.

\bibitem{chadwick2020vitruvio}
M. Chadwick, H. Kolvenbach, F. Dubois, H.~F. Lau, and M. Hutter.
\newblock Vitruvio: An open-source leg design optimization toolbox for walking
  robots.
\newblock {\em IEEE Robotics and Automation Letters}, {\bf 5}(4), 6318--6325
  (2020).
 
\bibitem{da20162d}
X. Da, O. Harib, R. Hartley, B. Griffin, and J.~W. Grizzle.
\newblock From {2D} design of underactuated bipedal gaits to {3D}
  implementation: Walking with speed tracking.
\newblock {\em IEEE Access}, {\bf 4}, 3469--3478 (2016).
 
\bibitem{grandia2019feedback}
R. Grandia, F. Farshidian, R. Ranftl, and M. Hutter.
\newblock Feedback {MPC} for torque-controlled legged robots.
\newblock In {\em 2019 IEEE/RSJ International Conference on Intelligent Robots
  and Systems (IROS)} 2019, pp. 4730--4737.

 
\bibitem{rathod2021nmpc}
N. Rathod, A. Bratta, M. Focchi, M. Zanon, O. Villarreal, C. Semini, and
  A. Bemporad.
\newblock Model predictive control with environment adaptation for legged
  locomotion.
\newblock {\em IEEE Access}, {\bf 9}, 145710--145727 (2021).
 
\bibitem{bratta2021nmpc}
A. Bratta, N. Rathod, M. Zanon, O. Villarreal, A. Bemporad, C. Semini, and
  M. Focchi.
\newblock Towards a nonlinear model predictive control for quadrupedal
  locomotion on rough terrain.
\newblock In {\em International Workshop on Intelligent Robotic Systems in
  Motion (IRIM)} 2021.
 
\bibitem{Le2024CIMPC}
S. Le~Cleac'h, T.~A. Howell, S. Yang, C.-Y. Lee, J. Zhang, A. Bishop,
  M. Schwager, and Z. Manchester.
\newblock Fast contact-implicit model predictive control.
\newblock {\em IEEE Transactions on Robotics}, {\bf 40}, 1617--1629 (2024).
 
\bibitem{winkler2018gait}
A.~W. Winkler, C.~D. Bellicoso, M. Hutter, and J. Buchli.
\newblock Gait and trajectory optimization for legged systems through
  phase-based end-effector parameterization.
\newblock {\em IEEE Robotics and Automation Letters}, {\bf 3}(3), 1560--1567
  (2018).
 
\bibitem{tobin2017domain}
J. Tobin, R. Fong, A. Ray, J. Schneider, W. Zaremba, and P. Abbeel.
\newblock Domain randomization for transferring deep neural networks from
  simulation to the real world.
\newblock In {\em 2017 IEEE/RSJ International Conference on Intelligent Robots
  and Systems (IROS)} 2017, pp. 23--30.

\bibitem{tan2018sim}
J. Tan, T. Zhang, E. Coumans, A. Iscen, Y. Bai, D. Hafner, S. Bohez, and
  V. Vanhoucke.
\newblock Sim-to-real: Learning agile locomotion for quadruped robots.
\newblock In {\em Robotics: Science and Systems XIV} 2018. Robotics: Science
  and Systems Foundation.
   
\bibitem{achiam2017constrained}
J. Achiam, D. Held, A. Tamar, and P. Abbeel.
\newblock Constrained policy optimization.
\newblock In D. Precup and Y.~W. Teh, editors, {\em Proceedings of the 34th
  International Conference on Machine Learning} 2017, volume~70 of {\em
  Proceedings of Machine Learning Research}, pp. 22--31. PMLR.
 
\bibitem{chow2018lyapunov}
Y. Chow, O. Nachum, A. Faust, E. Duenez-Guzman, and M. Ghavamzadeh 2019.
\newblock Lyapunov-based safe policy optimization for continuous control.
\newblock arXiv:1901.10031.
 
\bibitem{nachum2018data}
O. Nachum, S. Gu, H. Lee, and S. Levine.
\newblock Data-efficient hierarchical reinforcement learning.
\newblock In {\em Proceedings of the 32nd International Conference on Neural
  Information Processing Systems} 2018, pp. 3307--3317. Curran Associates Inc.

\bibitem{finn2017model}
C. Finn, P. Abbeel, and S. Levine.
\newblock Model-agnostic meta-learning for fast adaptation of deep networks.
\newblock In D. Precup and Y.~W. Teh, editors, {\em Proceedings of the 34th
  International Conference on Machine Learning} 2017, volume~70 of {\em
  Proceedings of Machine Learning Research}, pp. 1126--1135. PMLR.
 
\bibitem{yang2020data}
A. Yang, J. Hwangbo, C. Margolis, and M. Hutter.
\newblock Data-efficient reinforcement learning for legged robots.
\newblock In {\em Proceedings of the 34th Conference on Neural Information
  Processing Systems (NeurIPS)} 2020, pp. 1--12.

\bibitem{xie2021glide}
Z. Xie, X. Da, B. Babich, A. Garg, and M.~v. de~Panne.
\newblock {GLiDE}: Generalizable quadrupedal locomotion in diverse environments
  with a centroidal model.
\newblock In {\em Algorithmic Foundations of Robotics XV} 2023, pp. 523--539.
  Springer International Publishing.

\bibitem{chen2023learningtorque}
S. Chen, B. Zhang, M.~W. Mueller, A. Rai, and K. Sreenath 2023.
\newblock Learning torque control for quadrupedal locomotion.
\newblock arXiv:2203.05194.

\bibitem{10590932}
M. Weiss, A. Stirling, A. Pawluchin, D. Lehmann, Y. Hannemann, T. Seel, and
  I. Boblan.
\newblock Achieving velocity tracking despite model uncertainty for a quadruped
  robot with a {PD-ILC} controller.
\newblock In {\em 2024 European Control Conference (ECC)} 2024, pp. 134--140.
   
\bibitem{alqaham2024energy}
Y.~G. Alqaham, J. Cheng, and Z. Gan.
\newblock Energy-optimal asymmetrical gait selection for quadrupedal robots.
\newblock {\em IEEE Robotics and Automation Letters}, {\bf 9}(10), 8386--8393
  (2024).

\bibitem{gong2019feedback}
Y. Gong, R. Hartley, X. Da, A. Hereid, O. Harib, J.-K. Huang, and J. Grizzle.
\newblock Feedback control of a Cassie bipedal robot: Walking, standing, and
  riding a Segway.
\newblock In {\em 2019 American Control Conference (ACC)} 2019, pp. 4559--4566.

\bibitem{westervelt2007feedback}
E.~R. Westervelt, J.~W. Grizzle, C. Chevallereau, J.~H. Choi, and B. Morris
  2007.
\newblock {\em Feedback Control of Dynamic Bipedal Robot Locomotion}.
\newblock CRC Press.

\bibitem{hereid2017frost}
A. Hereid and A.~D. Ames.
\newblock {FROST*}: Fast robot optimization and simulation toolkit.
\newblock In {\em 2017 IEEE/RSJ International Conference on Intelligent Robots
  and Systems (IROS)} 2017, pp. 719--726.

\bibitem{raibert1986legged}
M.~H. Raibert 1986.
\newblock {\em Legged Robots That Balance}.
\newblock MIT Press.

\bibitem{tedrakeUnderactuated}
R. Tedrake 2022.
\newblock {\em Underactuated Robotics: Algorithms for Walking, Running,
  Swimming, Flying, and Manipulation}.
\newblock MIT Press.
\newblock \url{https://underactuated.mit.edu}.

\bibitem{6094484}
A. Chilian, H. Hirschm{\"u}ller, and M. G{\"o}rner.
\newblock Multisensor data fusion for robust pose estimation of a six-legged
  walking robot.
\newblock In {\em 2011 IEEE/RSJ International Conference on Intelligent Robots
  and Systems} 2011, pp. 2497--2504.
 
\bibitem{koenig2004design}
N. Koenig and A. Howard.
\newblock Design and use paradigms for {Gazebo}, an open-source multi-robot
  simulator.
\newblock In {\em 2004 IEEE/RSJ International Conference on Intelligent Robots
  and Systems (IROS)} 2004, volume~3, pp. 2149--2154.

\bibitem{Hof:1996}
A.~L. Hof.
\newblock Scaling gait data to body size.
\newblock {\em Gait and Posture}, {\bf 4}(3), 222--223 (1996).
  
\bibitem{bledt2018cheetah}
G. Bledt, B. Katz, J. Di~Carlo, P.~M. Wensing, and S. Kim.
\newblock {MIT} Cheetah 3: Design and control of a robust, dynamic quadruped
  robot.
\newblock In {\em 2018 IEEE/RSJ International Conference on Intelligent Robots
  and Systems (IROS)} 2018, pp. 2245--2252. IEEE.

\bibitem{mordatch2010robust}
I. Mordatch, M. de~Lasa, and A. Hertzmann.
\newblock Robust physics-based locomotion using low-dimensional planning.
\newblock In {\em ACM SIGGRAPH 2010 Papers} 2010, pp. 71:1--71:8. ACM.

\bibitem{nguyen2024mastering}
C. Nguyen, L. Bao, and Q. Nguyen 2024.
\newblock Mastering agile jumping skills from simple practices with iterative
  learning control.
\newblock arXiv:2408.02619.

\end{thebibliography}
\end{document}